\documentclass[journal]{IEEEtran}
\usepackage[linesnumbered,ruled]{algorithm2e}
\usepackage{graphicx,amssymb,mathrsfs,amsmath,array,color,amsthm}
\usepackage{subfigure}
\usepackage{multirow}
\usepackage{booktabs}
\usepackage{soul}
\usepackage{bm}
\usepackage{mathtools}
\usepackage{bbm}
\usepackage{makecell}
\usepackage{threeparttable}
\usepackage{algpseudocode}

\usepackage{cite}
\usepackage[colorlinks,urlcolor=blue,driverfallback=dvipdfm]{hyperref}
\usepackage{pifont}

\begin{document}
\title{FlexiMo: A Flexible Remote Sensing Foundation Model}

\author{Xuyang Li,
        Chenyu Li,
        Pedram~Ghamisi,~\IEEEmembership{Senior Member,~IEEE,}
        Danfeng Hong,~\IEEEmembership{Senior Member,~IEEE}
        
\thanks{This work was supported by the National Natural Science Foundation of China under Grant 42271350, by the International Partnership Program of the Chinese Academy of Sciences under Grant No.313GJHZ2023066FN, and the 3D-ABC project, funded through the Helmholtz Foundation Model Initiative.}
\thanks{X. Li and D. Hong are with the Aerospace Information Research Institute, Chinese Academy of Sciences, Beijing 100094, China, and the School of Electronic, Electrical and Communication Engineering, University of Chinese Academy of Sciences, Beijing 100049, China. (e-mail: lixuyang23@mails.ucas.ac.cn, hongdf@aircas.ac.cn)}
\thanks{C. Li is with the School of Mathematics and Statistics, Southeast University, Nanjing 211189, China. (e-mail: chenyuli.erh@gmail.com)}
\thanks{P. Ghamisi is with the Helmholtz-Zentrum Dresden-Rossendorf (HZDR), Helmholtz Institute Freiberg for Resource Technology, 09599 Freiberg, Germany, and also with the Lancaster Environment Centre, Lancaster University, LA1 4YR Lancaster, U.K. (e-mail: p.ghamisi@gmail.com)}
}

\markboth{}
{Shell \MakeLowercase{\textit{et al.}}: A Flexible Remote Sensing Foundation Model}

\maketitle
\begin{abstract}
The rapid expansion of multi-source satellite imagery is driving innovation in Earth observation, opening unprecedented opportunities for Remote Sensing Foundation Models to harness diverse data. However, many existing models remain constrained by fixed spatial resolutions and patch sizes, limiting their ability to fully exploit the heterogeneous spatial characteristics inherent in satellite imagery.
To address these challenges, we propose \textbf{FlexiMo}, a flexible remote sensing foundation model that endows the pre-trained model with the flexibility to adapt to arbitrary spatial resolutions. Central to FlexiMo is a spatial resolution-aware module that employs a parameter-free alignment embedding mechanism to dynamically recalibrate patch embeddings based on the input image’s resolution and dimensions. This design not only preserves critical token characteristics and ensures multi-scale feature fidelity but also enables efficient feature extraction without requiring modifications to the underlying network architecture.
In addition, FlexiMo incorporates a lightweight channel adaptation module that leverages prior spectral information from sensors. This mechanism allows the model to process images with varying numbers of channels while maintaining the data’s intrinsic physical properties.
Extensive experiments on diverse multimodal, multi-resolution, and multi-scale datasets demonstrate that FlexiMo significantly enhances model generalization and robustness. In particular, our method achieves outstanding performance across a range of downstream tasks, including scene classification, land cover classification, urban building segmentation, and cloud detection. By enabling parameter-efficient and physically consistent adaptation, FlexiMo paves the way for more adaptable and effective foundation models in real-world remote sensing applications.

\end{abstract}
\graphicspath{{figures/}}


\section{Introduction}
\IEEEPARstart{W}{ith} the explosive growth of multi-source satellite remote sensing data, Earth observation technologies have undergone revolutionary advancements, providing high-precision support for critical applications such as smart city management, ecological monitoring, and disaster emergency response~\cite{zhao2024artificial}. Simultaneously, breakthrough progress in general foundation models has opened new avenues for the intelligent interpretation of remote sensing data. Motivated by these developments, researchers have increasingly integrated domain expertise with large-scale remote sensing datasets to develop Remote Sensing Foundation Models (RSFMs) \cite{Li2024Interpretable}.
\begin{figure}[!t]
	\centering
	\includegraphics[width=1.0\linewidth]{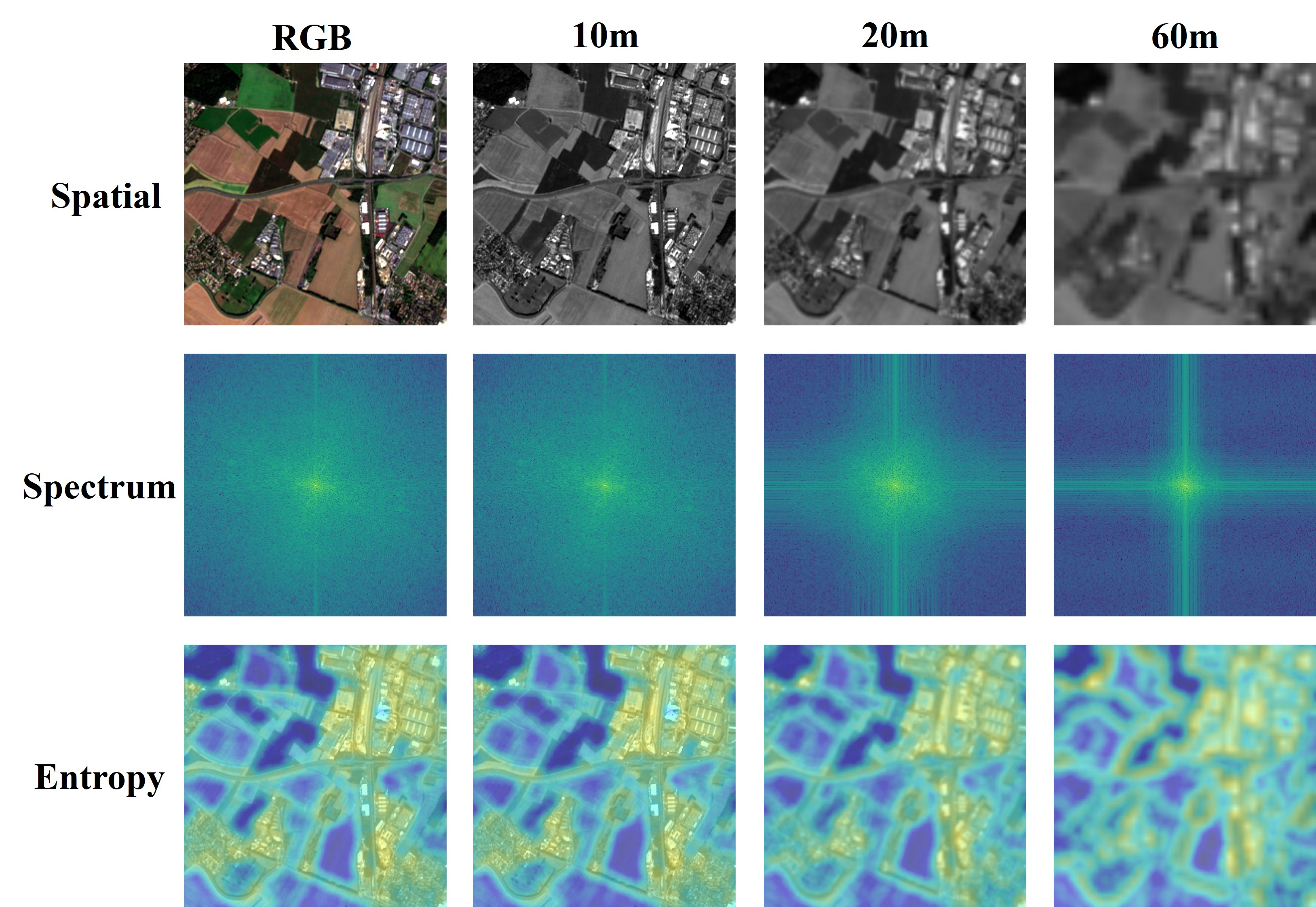}
	\caption{\textbf{Image Quality and Detail Across Various Spatial Resolutions:} Each column presents an image's spatial layout, its frequency spectrum, and the corresponding local information entropy heatmap.}
	\label{fig:motivation}
\end{figure}

RSFMs have evolved along a clear trajectory. Early models focused on single-modal processing of RGB imagery, later extending to multispectral and hyperspectral (MS/HS) data analysis~\cite{cong2022satmae,li2024s2mae}, and eventually developing into general frameworks supporting multi-source data fusion~\cite{hong2024multimodal,li2024seamomultiseasonalmultimodalremote}. This evolution has not only improved model input compatibility but also enhanced generalization in tasks such as image interpretation and scene understanding. Nonetheless, the inherent diversity in spatial resolutions and patch sizes across remote sensing imagery still poses significant challenges to existing architectures.

A central challenge lies in the fixed-size patch tokenization adopted by most models, particularly those based on the Vision Transformer (ViT) architecture. This approach gives rise to two major issues:
\begin{enumerate}
    \item \textbf{Rigid Tokenization Mechanism:} Partitioning images into fixed-size patches (e.g., $16\times16$ pixels) generates token sequences whose lengths vary with changes in input resolution. This variability disrupts the geometric consistency of positional encodings, thereby limiting the model's ability to accurately capture spatial details.
    \item \textbf{Multi-scale Perception Conflict:} The fixed patch size selected during pre-training forces a trade-off. In high-resolution images, small patches lead to significant computational overhead, while in low-resolution scenarios, larger patches risk discarding critical fine-grained details. Figure~\ref{fig:motivation} vividly illustrates this conflict.
\end{enumerate}

Furthermore, models originally tailored for high-resolution RGB data often struggle to process multispectral or low-resolution inputs. Similarly, spectral models with fixed bands (e.g., SpectralGPT~\cite{hong2024spectralgpt}) face challenges in adapting to diverse sensor specifications. Multimodal approaches such as RemoteCLIP~\cite{10504785} target image-text tasks, while solutions such as Geo-Chat~\cite{kuckreja2023geochat} empower large language models to interpret remote sensing imagery. Although these models perform well when the pre-training data distribution closely matches the downstream tasks, they are less effective under modality differences or distribution shifts.

Many studies have attempted to address spatial resolution variations. For instance, AnySat~\cite{astruc2024anysat} merges image tokens from sub-patches extracted at multiple resolutions, while ScaleMAE~\cite{reed2023scale} and SenPa-MAE~\cite{prexl2024senpa} incorporate resolution information at the data encoding stage. However, these models typically assume fixed spatial configurations during training and show suboptimal generalization under arbitrary conditions. In parallel, some research has explored channel adaptation mechanisms through techniques such as generating convolution kernels via central wavelength embedding~\cite{xiong2024neural}. Although effective for handling variable channel configurations, these approaches largely neglect the equally critical issue of flexible spatial resolution and patch size within the network architecture.

In remote sensing, the ability to flexibly adapt to diverse spatial resolutions is essential. Variations in sensor technology and imaging conditions result in data spanning a wide range of spatial scales. Rigid patch tokenization in existing models not only limits spatial adaptability but also constrains the design of network architectures that could inherently support flexible patch sizes. Enabling patch size flexibility directly within the model architecture allows for more efficient processing by adjusting the tokenization granularity to match the resolution of the input data. This adaptability not only optimizes computational efficiency in high-resolution scenarios but also ensures the preservation of critical fine-grained details in low-resolution images, ultimately enhancing the robustness of remote sensing applications.

Existing solutions based on parameter fine-tuning, post-training adjustments, or data preprocessing (e.g., resampling and merging) have significant drawbacks~\cite{li2024seamomultiseasonalmultimodalremote}. Data preprocessing can disturb the intrinsic physical properties of the original data, while extensive parameter fine-tuning demands substantial computational resources and risks overfitting, particularly in low-data scenarios such as change detection. An ideal solution should satisfy two conditions: (1) \textbf{Dimensional Independence:} supporting flexible usage across arbitrary spatial dimensions and resolutions within the model architecture; and (2) \textbf{Physical Consistency:} preserving the sensor’s inherent physical properties (e.g., the wavelength-channel correspondence).

To address these challenges, we propose FlexiMo, a flexible remote sensing foundation model. FlexiMo achieves dimensional independence via a spatial resolution-aware scaling module that dynamically adjusts the patch embedding structure according to the input image’s resolution and patch size. By inverting bilinear interpolation through a pseudo-inverse operation, this module preserves essential embedding properties, such as token norms, and enables consistent extraction of multi-scale, fine-grained features, without modifying the underlying architecture.

To ensure physical consistency, FlexiMo incorporates a lightweight channel adaptation mechanism based on central wavelength embedding. Each channel’s central wavelength is treated as prior knowledge and used to dynamically generate convolution kernels tailored to the spectral characteristics of the input. This enables the model to flexibly adapt to varying channel configurations while preserving the physical integrity of the sensor outputs.

Our main contributions are summarized as follows:
\begin{itemize}
    \item We We propose FlexiMo, the first flexible remote sensing foundation model that supports arbitrary input resolution, size, and channel count by dynamically adapting the tokenization process to spatial resolution, while preserving spectral fidelity through wavelength-aware design.
    \item We design a spatial resolution-aware scaling module based on pseudo-inverse bilinear interpolation for consistent multi-scale feature extraction, and a lightweight wavelength-guided channel adaptation mechanism for handling variable spectral configurations.
    \item We validate FlexiMo on diverse datasets across modalities, resolutions, and channel configurations (RGB, multispectral, SAR), achieving strong performance on image- and pixel-level tasks including scene classification, land cover segmentation, building extraction, and cloud detection.
\end{itemize}

\begin{figure*}[!t]
      \centering	   
      \includegraphics[width=1\textwidth]{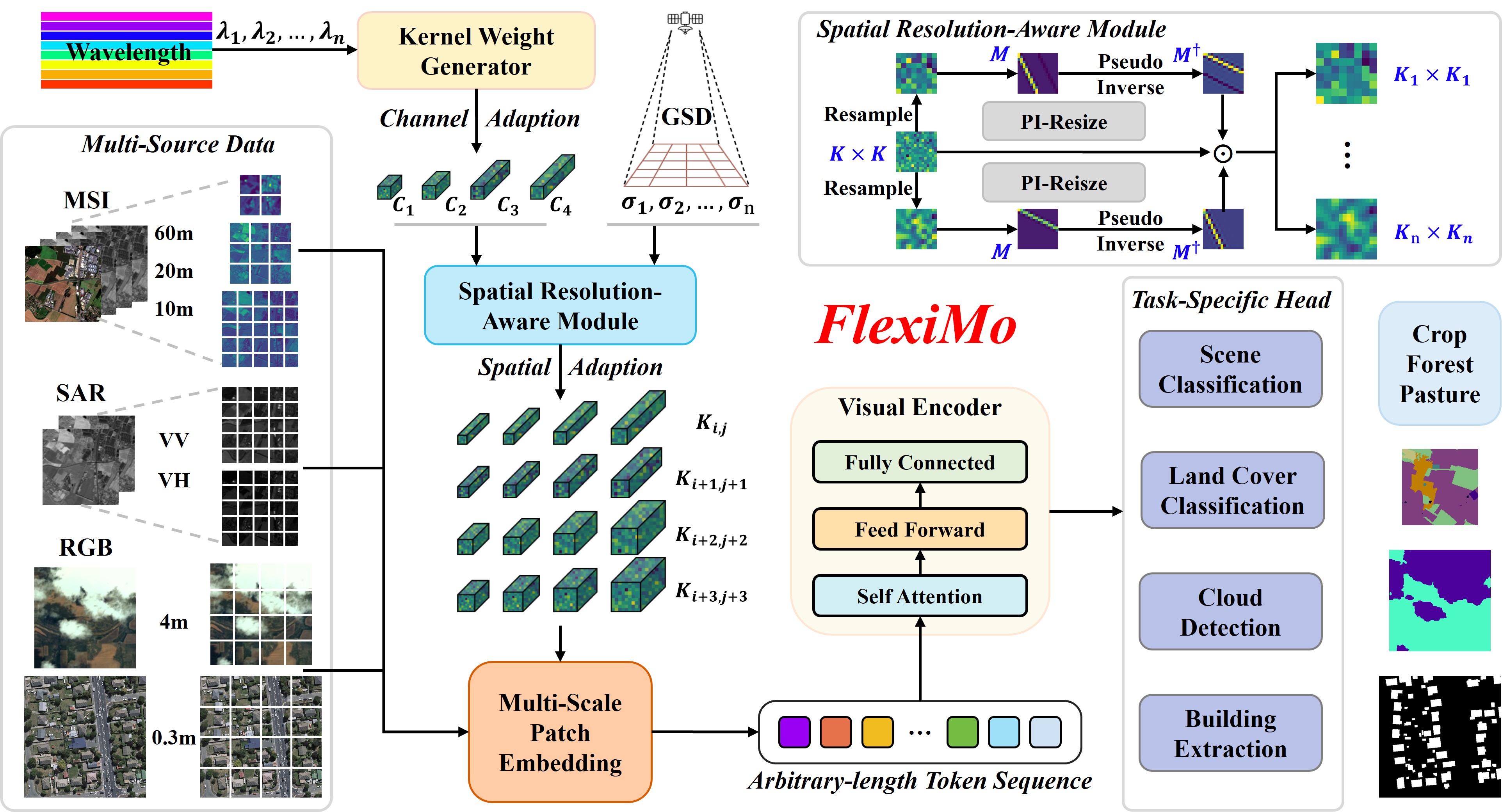}
      \caption{\textbf{An illustration of the proposed FlexiMo.} FlexiMo can adjust the patch embedding weights of a pretrained model based on the resolution and image
size of the input, enabling better adaptation to the data distribution of downstream tasks.}   
\label{fig:workflow}
\end{figure*}

\section{Related Work}
\subsection{Development of Vision Transformer}
Vision Transformer (ViT)~\cite{dosovitskiy2021an} has redefined image understanding by applying the transformer architecture to vision tasks. Starting from an input image 
\(
\mathbf{I} \in \mathbb{R}^{H \times W \times C},
\)
ViT first partitions \(\mathbf{I}\) into a grid of non-overlapping patches of size \(P \times P\). Each patch \(\mathbf{I}_p \in \mathbb{R}^{P \times P \times C}\) is then flattened into a vector and projected into a \(D\)-dimensional embedding via a learnable linear mapping:
\(
    \mathbf{Z}_p = \mathbf{W}_{\text{emb}}\, \mathrm{vec}\bigl(\mathbf{I}_p\bigr),
\)
with \(\mathbf{W}_{\text{emb}} \in \mathbb{R}^{(P^2 \cdot C) \times D}\).
To incorporate spatial information, learnable positional embeddings \(\mathbf{E}_{\text{pos}} \in \mathbb{R}^{N \times D}\) are added to the patch embeddings, where 
\(
N \;=\; \frac{H}{P} \times \frac{W}{P}
\)
denotes the total number of patches:
\(
    \tilde{\mathbf{Z}}_p = \mathbf{Z}_p + \mathbf{E}_{\text{pos}}(p).
\)
This enriched sequence \(\{\tilde{\mathbf{Z}}_p\}_{p=1}^N\), augmented by a dedicated learnable \([\text{CLS}]\) token for capturing global context, is subsequently processed by a standard transformer encoder, which consists of multi-head self-attention layers and feed-forward networks.
The core of this architecture is the self-attention mechanism, defined as:
\begin{equation}
\text{Attention}(Q, K, V) = \text{softmax}\!\left(\frac{QK^{T}}{\sqrt{d_k}}\right)V,
\end{equation}
where \(Q\), \(K\), and \(V\) denote the query, key, and value matrices, respectively, and \(d_k\) is the dimension of the key vectors. This operation enables the model to capture long-range dependencies and aggregate global contextual information effectively.

Transformer architectures have demonstrated outstanding performance in both image-level and pixel-level tasks. However, their rigid structure and massive computational requirements pose significant challenges~\cite{wang2025scaling}. To overcome these limitations, subsequent researchers have pursued several avenues to optimize network architectures based on the ViT. 
On one front, efforts have focused on integrating the hierarchical design of convolutional neural networks with transformers to enable the simultaneous learning of global and local features. For example, the Swin-Transformer~\cite{liu2021swin} introduces a hierarchical window-based attention mechanism that preserves the transformer’s ability to model long-range dependencies and, through window partitioning, enhances the focus on local details.
In parallel, considerable improvements have been achieved by refining the self-attention mechanism itself, as demonstrated by approaches like Linear Attention~\cite{katharopoulos2020transformers}. Furthermore, lightweight design has emerged as a crucial optimization direction. For example, EfficientViT~\cite{liu2023efficientvit} refines model architecture to decrease resource consumption while preserving performance.
Despite these advancements, the increasing complexity of image data has presented new challenges. During fine-tuning or inference, model performance often declines noticeably when the image size changes. To address this issue, various studies have been conducted. For example, FlexViT~\cite{beyer2023flexivit} introduces multiple patch embedding sizes during training to segment images and capture fine-grained information at different scales; ViTAR~\cite{fan2024vitar} accepts multi-size images and incorporates fuzzy position encoding to better adapt to varying image dimensions; and MSPE~\cite{liumspe} employs a training strategy with diverse image sizes and patch embeddings to jointly learn global and local features. In the field of remote sensing imagery, some studies~\cite{irvin2023usat,prexl2024senpa} have proposed spatial optimization strategies tailored to different image resolutions. However, these approaches have predominantly been tested under fixed, known conditions, and they have not demonstrated robust performance across arbitrary combinations of image sizes and patch embeddings.

\subsection{Remote Sensing Foundation Model}
Foundation models refer to those trained on extensive and diverse datasets, typically through large-scale self-supervised learning, which can then be adapted to a wide range of downstream tasks via fine-tuning or post-training~\cite{hong2024spectralgpt}. The rapid advancements of such models in natural language processing and computer vision have opened up new directions for remote sensing image processing. In the remote sensing domain, the development of RSFMs primarily follows two directions: Vision-only models and Vision-Language models.

Vision-only models mainly focus on leveraging self-supervised approaches, such as Masked Image Modeling (MIM) and Contrastive Learning (CL), to perform large-scale training on remote sensing data from various sensors, thereby learning robust representations that can be fine-tuned for downstream image understanding tasks. This approach has evolved from training on single-sensor data to incorporating multi-sensor datasets. For instance, SatMAE~\cite{cong2022satmae} and S2MAE~\cite{li2024s2mae} design Masked Autoencoder (MAE) strategies~\cite{he2022masked} that exploit the spectral characteristics of MS images. CROMA~\cite{fuller2023croma}, on the other hand, combines CL with MAE to train on both SAR and MS imagery. Moreover, some models enhance their representational power by integrating additional modalities such as geographical, audio, and temporal information~\cite{ayush2021geography,10655854}. Some models fine-tune generic architectures to tackle specific remote sensing tasks~\cite{zhang2024uv,li2025urbansam}, such as UrbanSAM~\cite{li2025urbansam}, which employs a multi-scale prompting mechanism to segment complex urban scenes. Models based on diffusion models or GANs that specialize in tasks like image generation, translation, and restoration~\cite{10663449}. For example, CRSDiff~\cite{10663449} utilizes ControlNet to generate remote sensing images from diverse input types. The development of Vision-Language models in remote sensing pursues two main objectives. One objective is to achieve alignment between text and remote sensing imagery to enhance image understanding, as demonstrated by models like RemoteCLIP~\cite{10504785}, which facilitate open-vocabulary and zero-shot tasks. The other objective focuses on enabling interaction between remote sensing images and large language models~\cite{kuckreja2023geochat}. Models such as Geochat~\cite{kuckreja2023geochat} are fine-tuned on remote sensing-specific instruction datasets, allowing them to capture multi-layered information inherent in remote sensing images.
Nevertheless, most of these models are pre-trained with fixed input configurations—using predetermined modalities, image sizes, and channels. Consequently, when downstream tasks involve data with varying characteristics, the models either cannot process the inputs properly or suffer a dramatic decline in performance, thus limiting the widespread application of remote sensing foundation models.

\section{METHODOLOGY}
\subsection{Overview of the Proposed FlexiMo}
\label{overview}
In this paper, we propose a flexible Remote Sensing foundation model, termed \textit{\textbf{FlexiMo}}, which enables unified perception across arbitrary spatial scales in remote sensing imagery. Most existing RSFMs are pre-trained on data with fixed patch sizes, fixed image dimensions, and fixed channels, posing significant challenges when these models are applied to downstream tasks with varying data characteristics. The primary objective of \textit{FlexiMo} is to achieve resolution adaptation of pre-trained models to multi-scale remote sensing images, thereby enhancing their generality and effectiveness across diverse remote sensing tasks.

Specifically, \textit{FlexiMo} takes three inputs: the remote sensing image \(\mathbf{I}\), the spatial patch size parameter \(P\) corresponding to the image resolution, and the corresponding electromagnetic wavelength parameters \(\{\lambda_i\}_{i=1}^{C}\). Figure~\ref{fig:workflow} shows the overall framework of our method. Initially, the framework employs a dynamic weight generator, composed of a series of Transformer blocks, to generate kernel patch embeddings for channel weights. Formally, given an input image \(\mathbf{I}_p\) and its associated wavelength parameters, the dynamic weight generator produces a patch embedding:
\(
\mathbf{Z}_p = f_{\text{dw}}\Bigl(\mathbf{I}_p, \{\lambda_i\}_{i=1}^{C}\Bigr),
\)
where \(f_{\text{dw}}(\cdot)\) denotes the function implemented by the dynamic weight generator. The resulting patch embeddings \(\{\mathbf{Z}_p\}\) are then fed into a Spatial Resolution-Aware Module, which aligns the embeddings in the spatial domain with minimal information loss based on the provided patch size \(P\):
\(
\tilde{\mathbf{Z}}_p = f_{\text{sr}}(\mathbf{Z}_p, P),
\)
ensuring that information from different spatial scales is effectively integrated. Subsequently, \(\tilde{\mathbf{Z}}_p\) is applied to the image \(\mathbf{I}_p\), which is patchified to form token sequences that serve as inputs to the pre-trained model. By dynamically adjusting the patch embeddings according to the input data's characteristics, \textit{FlexiMo} maintains efficient feature extraction while minimizing computational overhead. For images with low resolution or small dimensions, a smaller patch size is employed to capture finer details; conversely, for high-resolution or larger images, a larger patch size is used to preserve global context and reduce redundancy. Additionally, the channel dimensions of the patch embeddings are automatically generated based on the electromagnetic wavelength parameters, further enhancing the model's adaptive capabilities.

\subsection{Dynamic Channel Adaptation with Wavelength Embedding}
\begin{figure}[!t]
	\centering
	\includegraphics[width=0.95\linewidth]{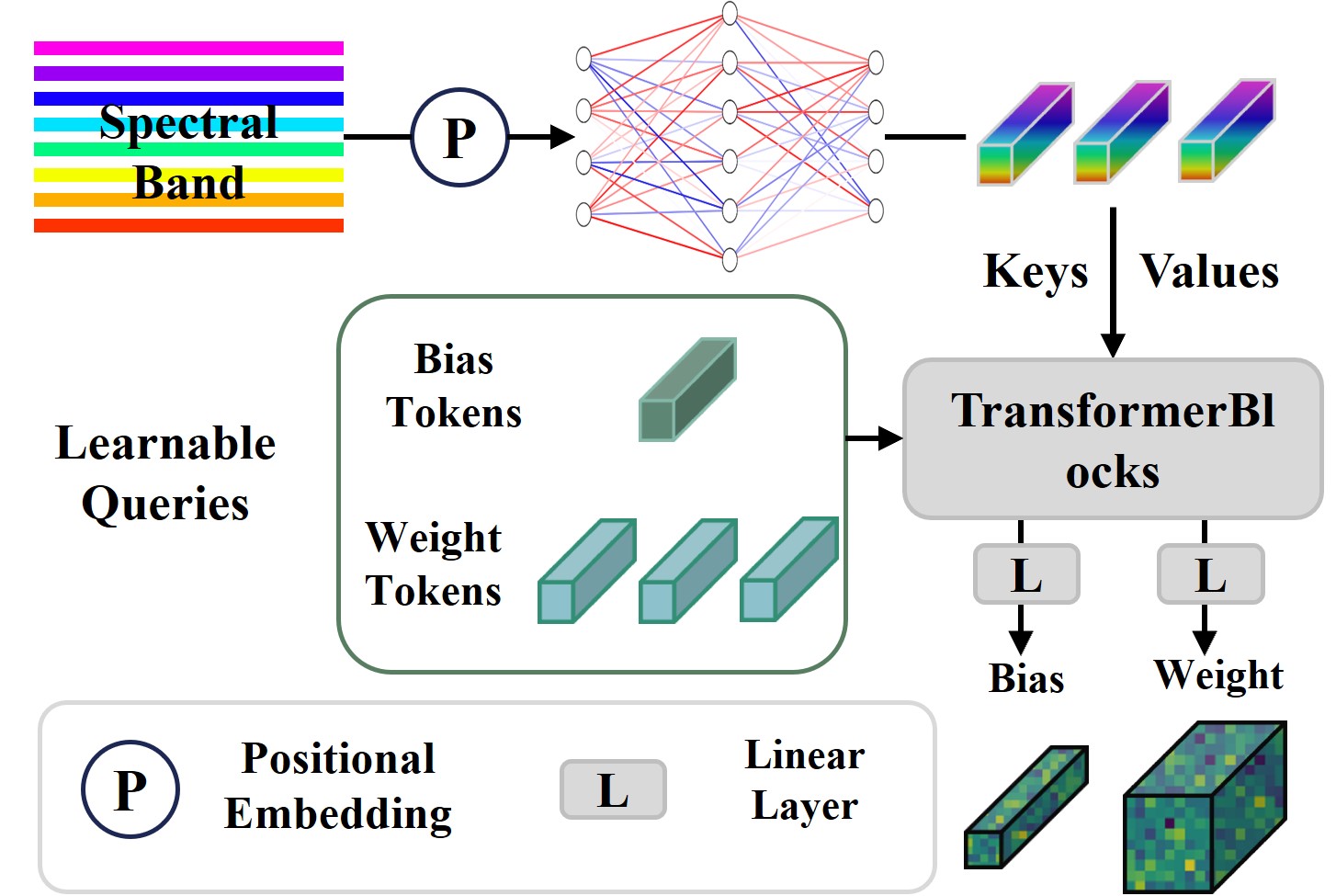}
	\caption{Illustration of the wavelength-guided dynamic weight generation process.}
	\label{fig:dofa}
\end{figure}
In conventional Vision Transformer architectures, the patch embedding process is typically performed using fixed convolution kernels, where an image \(\mathbf{I} \in \mathbb{R}^{C \times H \times W}\) is partitioned into patches using a predefined kernel. However, this approach assumes a fixed number of channels \(C\), which is not suitable for remote sensing images where each channel corresponds to a specific wavelength and the number of channels can vary. This limitation necessitates grouping or padding the channels to match the fixed kernel configuration, thereby disrupting the physical spectral properties of the data. Moreover, the fixed kernel size also restricts the model’s ability to adapt to varying spatial resolutions and patch sizes.

To address these challenges, we propose a flexible patch embedding mechanism that utilizes wavelength information as prior knowledge to generate dynamic convolution kernels. Rather than relying on fixed kernels, we treat the wavelength of each channel as an index to determine the channel configuration dynamically, guiding the model to generate convolution kernels specifically tailored to the spectral characteristics of each input image.

The first step in our approach is to encode the wavelength information, denoted as \(\lambda \in \mathbb{R}^{C}\), where each element represents the central wavelength of a channel. This wavelength vector is embedded into a higher-dimensional space using a positional encoding function \(\mathcal{P}(\cdot)\):
\(
\mathbf{E}_\lambda = \mathcal{P}(\lambda),
\)
where \(\mathbf{E}_\lambda \in \mathbb{R}^{C \times D}\) and \(D\) is the embedding dimension. Next, we apply a nonlinear transformation to the encoded wavelength information using a fully connected layer:
\(
\mathbf{E}'_\lambda = \phi\bigl(\mathbf{E}_\lambda\bigr),
\)
where \(\phi(\cdot)\) denotes the nonlinear activation function and \(\mathbf{E}'_\lambda \in \mathbb{R}^{C \times D}\) is the refined wavelength embedding.

At this stage, we introduce a lightweight Transformer-based module to dynamically generate the convolution kernels needed for patch tokenization. The key idea is that, unlike fixed architectures, the Transformer can process input sequences of arbitrary length. By encoding the wavelength information as a sequence, we leverage the Transformer’s self-attention mechanism to generate convolution kernels tailored to the specific spectral and spatial properties of the input image.

We begin by concatenating the refined wavelength embedding \(\mathbf{E}'_\lambda\) with learnable weight query tokens \(\mathbf{Q}_w \in \mathbb{R}^{L \times D}\) and a learnable bias query token \(\mathbf{Q}_b \in \mathbb{R}^{1 \times D}\). This concatenated sequence is processed by the Transformer encoder, and we extract the outputs corresponding to the query tokens:
\(
\begin{aligned}
\mathbf{Z}_w &\in \mathbb{R}^{L \times D}, \mathbf{Z}_b &\in \mathbb{R}^{D},
\end{aligned}
\)
such that
\begin{equation}
[\mathbf{Z}_w; \mathbf{Z}_b] = \mathcal{T}\Bigl(\operatorname{Concat}(\mathbf{Q}_w, \mathbf{E}'_\lambda, \mathbf{Q}_b)\Bigr).
\end{equation}
The self-attention mechanism allows the Transformer to generate meaningful representations that map to the required convolution parameters. These outputs are then passed through fully connected layers \(g_w(\cdot)\) and \(g_b(\cdot)\) to produce the final convolution weights and biases:
\begin{equation}
\mathbf{W}_{\text{dyn}} = g_w\Bigl(\mathbf{Z}_w + \mathbf{E}'_\lambda\Bigr) \in \mathbb{R}^{C \times P^2 \times D},
\end{equation}
\begin{equation}
\mathbf{B}_{\text{dyn}} = g_b\Bigl(\mathbf{Z}_b\Bigr) \in \mathbb{R}^{C \times D}.
\end{equation}
These weights and biases are reshaped to form the convolution kernel:
\begin{equation}
\mathbf{K}_{\text{dyn}} = \operatorname{Reshape}\bigl(\mathbf{W}_{\text{dyn}}, [D, C, P, P]\bigr),
\end{equation}
which is then applied to the input image \(\mathbf{I}\) for patch embedding:
\begin{equation}
\mathbf{Z}_p = \operatorname{Conv}\bigl(\mathbf{I}, \mathbf{K}_{\text{dyn}}, \mathbf{B}_{\text{dyn}}\bigr).
\end{equation}

This formulation decouples the patch embedding process from a fixed spectral configuration. By generating convolution kernels based on the central wavelengths of the channels, our approach enables the model to learn channel-specific representations on the fly. Figure~\ref{fig:dofa} illustrates the detailed operations involved in dynamically generating convolutional weights and biases. It should be noted that this dynamic weight generator also has certain limitations. On one hand, it requires extensive pre-training on a large-scale dataset. If the pre-training phase is skipped and the model is directly fine-tuned on a small-scale dataset, the lack of sufficient prior knowledge may lead to performance inferior to that of a standard Transformer model. On the other hand, although the dynamic weight generation mechanism addresses channel flexibility, its reliance on fixed image sizes and fixed patch sizes still poses challenges for spatial adaptability. In the following sections, we present solutions to these issues during fine-tuning.

\subsection{Spatial Resolution-Aware Module with Parameter-Free Alignment Embedding}

Remote sensing images exhibit varying spatial resolutions depending on the sensor type. Unlike natural images (e.g., those from ImageNet), whose resolution is defined by pixel dimensions, the spatial resolution of remote sensing images is generally quantified by the Ground Sample Distance (GSD)—the actual distance between the centers of adjacent pixels on the ground. This difference implies that the perceptual geometry embedded in the raw input plays a crucial role in downstream tasks.

In pre-trained foundation models with ViT as the backbone, the input image is first partitioned into a series of non-overlapping patches of fixed size \(P \times P\). Each patch \(\mathbf{I}_p \in \mathbb{R}^{P \times P \times C}\) is then flattened and projected into a \(D\)-dimensional embedding space via a learnable linear mapping:
\(
\mathbf{Z}_p = \mathbf{W}_{\text{emb}}\,\operatorname{vec}\bigl(\mathbf{I}_p\bigr).
\)
However, this fixed patch embedding structure inherently locks the spatial resolution perception during tokenization, even if additional resolution encoding or other prior information is later incorporated~\cite{reed2023scale,prexl2024senpa}. Furthermore, simply applying bilinear interpolation to both the patch and the embedding weights tends to cause dramatic changes in the token norm~\cite{moore1920reciprocal}. To mitigate this issue, it is essential to adjust the model's patch embedding to be spatial resolution-aware while preserving the intrinsic information. We propose a \emph{Spatial Resolution-Aware Module} that is plug-and-play and parameter-free. The key idea is to employ a pseudo-inverse matrix transformation (PI-Resize) to “invert” the distortion introduced by standard bilinear interpolation.

Concretely, let the original patch embedding kernel be denoted by
\(
\mathbf{K}_{\text{old}} \in \mathbb{R}^{H \times W \times D \times O},
\)
where \(H\) and \(W\) are the spatial dimensions of the kernel, \(D\) is the input feature dimension, and \(O\) is the output dimension. Our objective is to compute a new kernel
\(
\mathbf{K}_{\text{new}} \in \mathbb{R}^{H' \times W' \times D \times O},
\)
which is aligned with the target spatial resolution without losing any information.

First, we define a bilinear interpolation operator \(\mathcal{R}\) that maps a 2D array of size \((H, W)\) to one of size \((H', W')\). In linear algebra form, this operation can be expressed as:
\begin{equation}
\operatorname{resize}(o) = M\,\operatorname{vec}(o),
\end{equation}
where \(o \in \mathbb{R}^{H \times W}\) and \(M \in \mathbb{R}^{(H'W') \times (HW)}\) is the corresponding bilinear interpolation matrix. Directly applying \(M\) to both the input and the embedding weights alters the overall token norm. We hypothesize that such norm distortion contributes to the observed inflexibility, and that an inverse operation is needed to preserve the original embedding properties.
To this end, we construct the matrix \(M\) by applying \(\mathcal{R}\) to the standard basis vectors in \(\mathbb{R}^{H \times W}\). Assuming \(M^\top M\) is invertible, we compute the Moore–Penrose pseudo-inverse:
\begin{equation}
M^\dagger = \bigl(M^\top M\bigr)^{-1}M^\top.
\end{equation}
This pseudo-inverse \(M^\dagger\) serves as an “inverse” of the bilinear interpolation, restoring the token norm to its original scale provided no information is lost. For each channel slice \(\mathbf{K}_{\text{old}}^c \in \mathbb{R}^{H \times W}\) (with \(c = 1, 2, \dots, D \times O\)), we first flatten it to a vector \(\operatorname{vec}(\mathbf{K}_{\text{old}}^c) \in \mathbb{R}^{HW}\), then apply the pseudo-inverse transformation:
\begin{equation}
\mathbf{K}_{\text{new}}^c = \operatorname{reshape}\Bigl(M^\dagger \cdot \operatorname{vec}(\mathbf{K}_{\text{old}}^c)\Bigr) \in \mathbb{R}^{H' \times W'}.
\end{equation}
Stacking these slices across all channels yields the final resized kernel:
\[
\mathbf{K}_{\text{new}} \in \mathbb{R}^{H' \times W' \times D \times O}.
\]

From a theoretical standpoint, this procedure can be viewed as solving the following optimization problem~\cite{beyer2023flexivit}:
\begin{equation}
\hat{\omega} \in \arg\min_{\hat{\omega}} \, \mathbb{E}_{x \sim X}\Bigl[\Bigl\| h(x, \omega_i) - h(Mx, \hat{\omega}_i) \Bigr\|^2\Bigr],
\end{equation}
where \(X\) represents the patch distribution. In the case of upsampling (i.e., when \(H'W' \ge HW\)), one can show that setting \(\hat{\omega} = P\,\omega\), with
\begin{equation}
P = M(M^\top M)^{-1} = \bigl(M^\top\bigr)^+,
\end{equation}
precisely recovers the original token embedding, i.e.,
\begin{equation}
h\bigl(Mx, \hat{\omega}_i\bigr) = h(x, \omega_i).
\end{equation}
For downsampling scenarios (when \(H'W' < HW\)), although the optimal solution depends on the underlying patch distribution (e.g., assuming \(X \sim \mathcal{N}(0,I)\), the pseudo-inverse still yields the best approximation), the PI-Resize operation provides a principled mechanism to minimize information loss.

By employing the Spatial Resolution-Aware Module, we can overcome the rigid tokenization mechanism of ViT models while simultaneously capturing multi-scale information in remote sensing images. For images of arbitrary resolution, the module dynamically adjusts the patch size to the most appropriate value, enabling effective tokenization and feature extraction. Although PI-Resize was first applied to patch embedding scaling in FlexViT~\cite{beyer2023flexivit}, our Spatial Resolution-Aware Module differs from FlexViT in three main aspects. First, we design our model with the spatial resolution of remote sensing images in mind. Second, while FlexViT merely proposes a neural network framework, we integrate PI-Resize within the context of enabling pre-trained foundation models to adapt to downstream tasks, thereby advancing spatial adaptive fine-tuning. Third, our overall framework is combined with channel adaptation, simultaneously addressing the adaptation of both arbitrary channels and spatial data.

\section{Experiments}
In this section, we present a comprehensive set of experiments to validate the effectiveness of our proposed method. We begin by outlining the implementation details of our experiments. Next, we report the results obtained by fine-tuning the pre-trained model on various downstream tasks, including medium-resolution MSI scene classification, medium-resolution SAR scene classification, medium-resolution MSI land cover classification, high-resolution RGB building detection, and high-resolution RGB cloud detection. Additionally, extensive ablation studies are conducted to further substantiate the efficacy of our approach.

\subsection{Implementation Details}
\subsubsection{Experimental Setup and Dataset Information}
In all our experiments, we utilize NVIDIA GeForce RTX 4090 GPUs and AMD EPYC
7Y83 CPUs. Specifically, image-level tasks are executed on 2 GPUs, while pixel-level tasks run on 4 GPUs. For image-level evaluation, we employ the EuroSAT~\cite{helber2019eurosat} and EuroSAT-SAR~\cite{wang2023feature} datasets, and for pixel-level evaluation, we use the SegMunich~\cite{hong2024spectralgpt}, WHU-Building~\cite{ji2018fully}, and GF12MS-WHU~\cite{zhu2024transferring} datasets. The information for all datasets is listed in Table~\ref{tab:dataset_info}. 
\subsubsection{Model Fine-Tuning and Task-Specific Adaptation}
Our approach primarily leverages the pre-trained weights from the DOFA model~\cite{xiong2024neural}, specifically its ViT-Base variant, which was originally pre-trained with image sizes of 224×224 and patch sizes of 16×16. Therefore, when our model is fine-tuned with a 224$\times$224 image size and a patch size of 16, it essentially degrades into the original DOFA model, yielding experimental results identical to those of DOFA. During experiments, the input wavelengths are set according to the central wavelengths of the different sensor types used in~\cite{xiong2024neural}. For each downstream task, we fully fine-tune the pre-trained model over all parameters using our proposed method, adapting to various channel configurations and patch sizes. For image-level tasks, we extract the \textit{[CLS]} token from the encoder of the pre-trained model and then connect it to a classification head for prediction; for pixel-level tasks, after feature extraction by the pre-trained model, we attach a UperNet~\cite{xiao2018unified} segmentation head.
\subsubsection{Model Evaluation Metrics}
To evaluate the performance of the proposed models, we adopt different metrics depending on the task.
For single-label scene classification tasks, we use Overall Accuracy (OA) as the evaluation metric,
while for pixel-level tasks such as segmentation, we use OA, precision, recall, and Intersection over Union (IoU).
The corresponding formulas are given by:
\begin{equation}
\text{OA} = \frac{TP + TN}{TP + TN + FP + FN}
\label{eq:oa}
\end{equation}

\begin{equation}
\text{Recall} = \frac{TP}{TP + FN}
\label{eq:recall}
\end{equation}

\begin{equation}
\text{Precision} = \frac{TP}{TP + FP}
\label{eq:precision}
\end{equation}

\begin{equation}
\text{IoU} = \frac{TP}{TP + FP + FN}
\label{eq:miou}
\end{equation}

where $TP$, $TN$, $FP$, and $FN$ denote true positives, true negatives, false positives, and false negatives, respectively.

\begin{table*}[!t]
    \centering
    \caption{Dataset information used in this paper. MSI represents multispectral images, SAR represents Synthetic Aperture Radar images, RGB represents high-resolution RGB images, and PAN represents panchromatic images.}
    \begin{tabular}{c c c c c c c c}
        \toprule
        Dataset & Sensors & Resolution & Modalities & Data Scale & Image Size & Tasks & Task Type \\
        \midrule
        EuroSAT~\cite{helber2019eurosat}       & Sentinel-2   & 10m,20m,60m   & MSI(13)          & 27,000 & 64x64   & \makecell{Scene \\Classification} & \multirow{3}{*}{Image Level} \\
        EuroSAT-SAR~\cite{wang2023feature}   & Sentinel-1   & 10m           & SAR(2)           & 27,000 & 64x64   & \makecell{Scene \\Classification} & \\
        \hline
        SegMunich~\cite{hong2024spectralgpt}     & Sentinel-2   & 10m,20m       & MSI(10)          & 49,248 & 128x128 & \makecell{Land Cover\\ Classification} & \multirow{5}{*}{Pixel Level} \\
        WHU-Building~\cite{ji2018fully}  & \makecell{Aerial,QuickBird,\\Worldview,IKONOS,ZY-3} & 0.3m to 2.5m & RGB(3) & 8393   & 512x512 & \makecell{Building \\Extraction}      & \\
        GF12MS-WHU~\cite{zhu2024transferring}    & GaoFen-1,GaoFen-2     & 2m,4m,8m      & MSI(4),PAN(1)     & 21,907  & 256x256 & \makecell{Cloud \\Detection}       & \\
        \bottomrule
    \end{tabular}
    \label{tab:dataset_info}
\end{table*}
\subsection{Performance on Image-level Tasks}
\subsubsection{Performance on EuroSAT Dataset}
The EuroSAT dataset~\cite{helber2019eurosat} comprises 27,000 Sentinel-2 satellite images collected from 34 European countries, offering a wealth of geographical and environmental information. The images exhibit diverse spatial resolutions across different spectral bands—specifically, 10 m, 20 m, and 60 m. They are meticulously categorized into 10 scene classes, including industrial buildings, residential areas, crops, permanent crops, rivers, oceans/lakes, herbaceous vegetation, highways, pastures, and forests, with each class containing between 2,000 and 3,000 labeled images. Each image in the dataset has been standardized to a resolution of \(64 \times 64\) pixels and covers 13 spectral bands, ranging from the visible spectrum to the shortwave infrared. To optimize data quality and ensure a fair evaluation of model performance, we followed the approach presented in~\cite{hong2024spectralgpt} by excluding the cirrus band (B10) and using only the remaining 12 bands for analysis and modeling. We adopted the same dataset split as in \cite{manas2021seasonal}, utilizing 16,200 images for training and 5,400 images for validation. All experimental results are reported on the validation set.

Table~\ref{tab:EuroSAT} shows the fine-tuning results on EuroSAT for our model compared with current mainstream RSFMs. When the image size used for fine-tuning matches that of the pre-training data, the performance is generally optimal~\cite{dosovitskiy2021an}. Therefore, for each model, we set the image size to be consistent with that of its pre-training dataset. The results demonstrate that our model, with a patch size of 8 and an image size of $224\times224$, achieves state-of-the-art (SOTA) performance. In our experiments, we set the batch size to 256, the base learning rate to \(5 \times 10^{-4}\), and trained for 200 epochs. The data augmentation strategy follows~\cite{hong2024spectralgpt}, including mixup with a coefficient of 0.8, cutmix with a coefficient of 1.0, and a drop path rate of 0.2. Notably, our model is the only one that allows for an adjustable patch size; its performance remains essentially stable across variations in image size and patch size, achieving an accuracy of 99.43 even with extremely small image sizes (e.g., $56\times56$).
\begin{table}[!t]
  \centering
  \setlength{\tabcolsep}{2pt} 
  \caption{Fine-Tuning results of Different models on the EuroSAT dataset. ViT-B denotes the ViT-Base model, while ViT-L denotes the ViT-Large model. Best and second-best results are in red and blue, respectively.}
  \label{tab:EuroSAT}
  \begin{tabular}{ccccc}
    \toprule
    Method & Backbone & Image Size & Patch Size & OA \\
    \midrule
    \midrule
    SeCo~\cite{manas2021seasonal}       & ResNet50  & 128$\times$128    & -      & 97.23 \\
    \midrule
    ViT-22k~\cite{dosovitskiy2021an}    & ViT-B     & 128$\times$128    & 16$\times$16  & 98.91 \\
    \midrule
    \multirow{2}{*}{SatMAE~\cite{cong2022satmae}} & ViT-B    & \multirow{2}{*}{120$\times$120}    & \multirow{2}{*}{8$\times$8}   & 99.20 \\
                           & ViT-L    &                                     &                          & 99.35 \\
    \midrule
    \multirow{2}{*}{SpectralGPT~\cite{hong2024spectralgpt}} & ViT-B   & \multirow{2}{*}{128$\times$128}    & \multirow{2}{*}{8$\times$8}   & 99.21 \\
                                 & ViT-L   &                                     &                          & 99.31 \\
    \midrule
    CROMA~\cite{fuller2023croma}     & ViT-B    & 120$\times$120    & 8$\times$8   & 99.22 \\
    \midrule
    \multirow{3}{*}{DOFA~\cite{xiong2024neural}}  & \multirow{3}{*}{ViT-B}    & 112$\times$112    & \multirow{3}{*}{16$\times$16}  & 99.27 \\
                           &                           & 128$\times$128    &                          & 99.17 \\
                           &                           & 224$\times$224    &                          & 99.39 \\
    \midrule
    SeaMo~\cite{li2024seamomultiseasonalmultimodalremote}     & ViT-B    & 128$\times$128    & 8$\times$8    & 99.37 \\
    \midrule
    \multirow{5}{*}{\makecell{\textbf{FlexiMo}\\\textbf{(Ours)}}} & \multirow{5}{*}{ViT-B}   & 56$\times$56   & 4$\times$4   & \textbf{\textcolor{blue}{\text{99.43}}} \\
                                &                           & 112$\times$112   & 8$\times$8   & 99.37 \\
                                &                           & 128$\times$128   & 8$\times$8   & 99.35 \\
                                &                           & 224$\times$224   & 16$\times$16 & 99.39 \\
                                &                           & 224$\times$224   & 8$\times$8   & \textbf{\textcolor{red}{\text{99.44}}} \\
    \bottomrule
  \end{tabular}
\end{table}
\subsubsection{Performance on EuroSAT-SAR Dataset}
EuroSAT-SAR~\cite{wang2023feature} is the SAR version of the EuroSAT dataset. The images are derived from Sentinel-1's products and include two modes: VV (vertical transmit, vertical receive) and VH (vertical transmit, horizontal receive), resulting in two-channel data. This dataset is geo-referenced to correspond with the EuroSAT image, with each image standardized to a size of \(64 \times 64\) pixels and covering the same 10 classes as EuroSAT, totaling 27,000 images. We adopt the same dataset split as EuroSAT, using only the train and val sets.

Table~\ref{tab:EuroSAT-SAR} presents the results of our model compared with baseline models on the EuroSAT-SAR dataset. The training parameters are identical to those used for the EuroSAT multispectral dataset. The experimental results demonstrate that our model maintains excellent spatial scale awareness even on SAR data. Furthermore, as the patch size is reduced, the model's performance improves significantly. For instance, with an image size of $128\times128$, reducing the patch size from $16\times16$ to $8\times8$ results in a 3.32\% increase in accuracy; similarly, with an image size of $224\times224$, modifying the patch size yields a 1.37\% improvement in accuracy.
\begin{table}[!t]
  \centering
  \setlength{\tabcolsep}{2pt} 
  \caption{Fine-Tuning results of different models on the EuroSAT-SAR dataset. ViT-S denotes the ViT-Small model, and ViT-B denotes the ViT-Base model. Best and second-best results are in red and blue, respectively.}
  \label{tab:EuroSAT-SAR}
  \begin{tabular}{ccccc}
    \toprule
    Method        & Backbone & Image Size         & Patch Size         & OA \\
    \midrule
    \midrule
    DINO-MM~\cite{wang2022self}       & ViT-S    & 128$\times$128      &\multirow{3}{*}{8$\times$8}         & 85.43 \\
    
    FGMAE~\cite{wang2023feature}         & ViT-S    & 120$\times$120      &         & 85.90 \\
    
    CROMA~\cite{fuller2023croma}         & ViT-B    & 120$\times$120      &          & 88.42 \\
    \midrule
    DOFA~\cite{xiong2024neural}          & ViT-B    & 224$\times$224      & 16$\times$16       & 88.59 \\
    \midrule
    SeaMo~\cite{li2024seamomultiseasonalmultimodalremote}         & ViT-B    & 128$\times$128      & 8$\times$8         & 89.69 \\
    \midrule
    \multirow{4}{*}{\makecell{\textbf{FlexiMo}\\\textbf{(Ours)}}} & \multirow{4}{*}{ViT-B}  & 128$\times$128  & 16$\times$16   & 86.41 \\
                                &                        & 128$\times$128  & 8$\times$8     & \textbf{\textcolor{blue}{\text{89.73}}} \\
                                &                        & 224$\times$224  & 16$\times$16   & 88.59 \\
                                &                        & 224$\times$224  & 8$\times$8     & \textbf{\textcolor{red}{\text{89.96}}} \\
    \bottomrule
  \end{tabular}
\end{table}
\begin{figure*}[!t]
	\centering
	\includegraphics[width=1.0\linewidth]{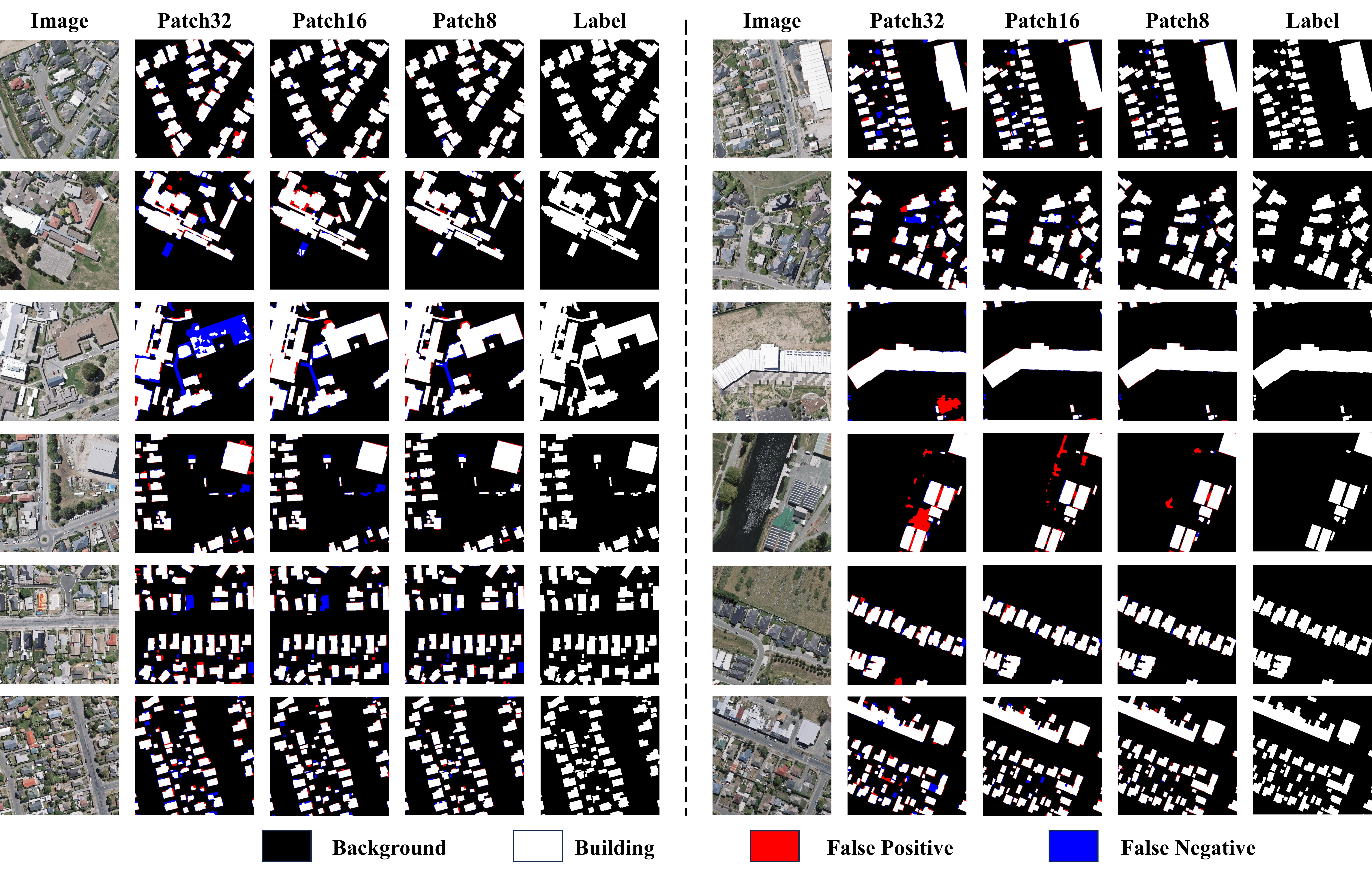} 
	\caption{Comparison of segmentation visualizations from models employing different patch sizes on the WHU-Building dataset. As the patch size decreases, the model extracts building edges more effectively.}
	\label{fig:building}
\end{figure*}
\subsection{Performance on Pixel-level Tasks}
\subsubsection{Performance on SegMunich Dataset}
\label{Performance_on_SegMunich Dataset}
The SegMunich dataset~\cite{hong2024spectralgpt} is constructed from Sentinel-2 MS satellite imagery, with a focus on capturing remote sensing data for Munich and its surrounding regions. It covers 13 land use and land cover (LULC) classes and provides detailed annotated masks for land cover classification tasks. In terms of scale, SegMunich comprises 39,402 training images and 9,846 validation images, with each image resized to \(128 \times 128\) pixels.
Only the 10-meter and 20-meter resolution bands (a total of 10 bands) from Sentinel-2 imagery are utilized, while the 60-meter resolution bands are discarded to reduce the adverse impact of low-resolution data on classification accuracy. Additionally, the 20-meter resolution bands are upsampled to align with the 10-meter resolution bands.

For medium-resolution remote sensing semantic segmentation tasks, we believe that resizing the labels to a different image size can adversely affect the segmentation accuracy evaluation. Therefore, we maintain the original image size of $128 \times 128$ pixels for all models. The training configuration is set with a batch size of 128, a learning rate of \(6 \times 10^{-5}\), and 100 epochs of training, with data augmentation limited to simple horizontal and vertical flips. Table~\ref{tab:SegMunich} presents the experimental results on the SegMunich dataset for our model and the comparison models. Notably, our proposed method achieves SOTA performance, outperforming the second-best method by 1.4 in terms of mIoU, a substantial improvement for semantic segmentation tasks.
\begin{table}[!t]
  \centering
  \setlength{\tabcolsep}{4pt} 
  \caption{Fine-Tuning results of different models on the SegMunich Dataset. ViT-B denotes the ViT-Base model. Best and second-best results are in red and blue, respectively.}
  \label{tab:SegMunich}
  \begin{tabular}{@{}lccc@{}}
    \toprule
    Method & Backbone & OA  & mIoU \\
    \midrule
    GASSL~\cite{ayush2021geography} & ResNet50 & 80.1 & 45.7 \\
    SeCo~\cite{manas2021seasonal}   & ResNet50 & 80.3 & 45.9 \\
    SatMAE~\cite{cong2022satmae}      & ViT-B    & 81.5 & 48.7 \\
    SpectralGPT~\cite{hong2024spectralgpt} & ViT-B & 82.7 & 51.0 \\
    DOFA~\cite{xiong2024neural}      & ViT-B    & 80.8 & 47.6 \\
    CROMA~\cite{fuller2023croma}     & ViT-B    & 82.9 & 51.1 \\
    SeaMo~\cite{li2024seamomultiseasonalmultimodalremote} & ViT-B & 82.8 & \textbf{\textcolor{blue}{\text{51.3}}} \\
    \midrule
    \textbf{FlexiMo (Ours)} & ViT-B & 83.4 & \textbf{\textcolor{red}{\text{52.7}}} \\
    \bottomrule
  \end{tabular}
\end{table}

\subsubsection{Performance on WHU-Building Dataset}
The WHU Building Dataset~\cite{ji2018fully} is an aerial and satellite imagery dataset designed for building extraction. The global cities satellite dataset portion contains images from cities around the world, sourced from various remote sensing platforms such as QuickBird, WorldView series, IKONOS, and ZY-3. It comprises 204 images, each with a resolution of $512 \times 512$ pixels and spatial resolutions ranging from 0.3\,m to 2.5\,m. In the aerial imagery subset, the original images are obtained from the New Zealand Land Information Services website. Researchers down-sampled most of these aerial images, which include approximately 187,000 buildings, to a ground resolution of 0.3\,m, and then cropped them into 8,189 tiles of $512 \times 512$ pixels. The ready-to-use samples are divided into three parts: a training set (130,500 buildings, 4,736 images), a validation set (14,500 buildings, 1,036 images), and a test set (42,000 buildings, 2,416 images). In this study, we employ the aerial imagery subset to evaluate the performance of our proposed method on high-resolution image segmentation.
\begin{figure*}[!t]
	\centering
	\includegraphics[width=1.0\linewidth]{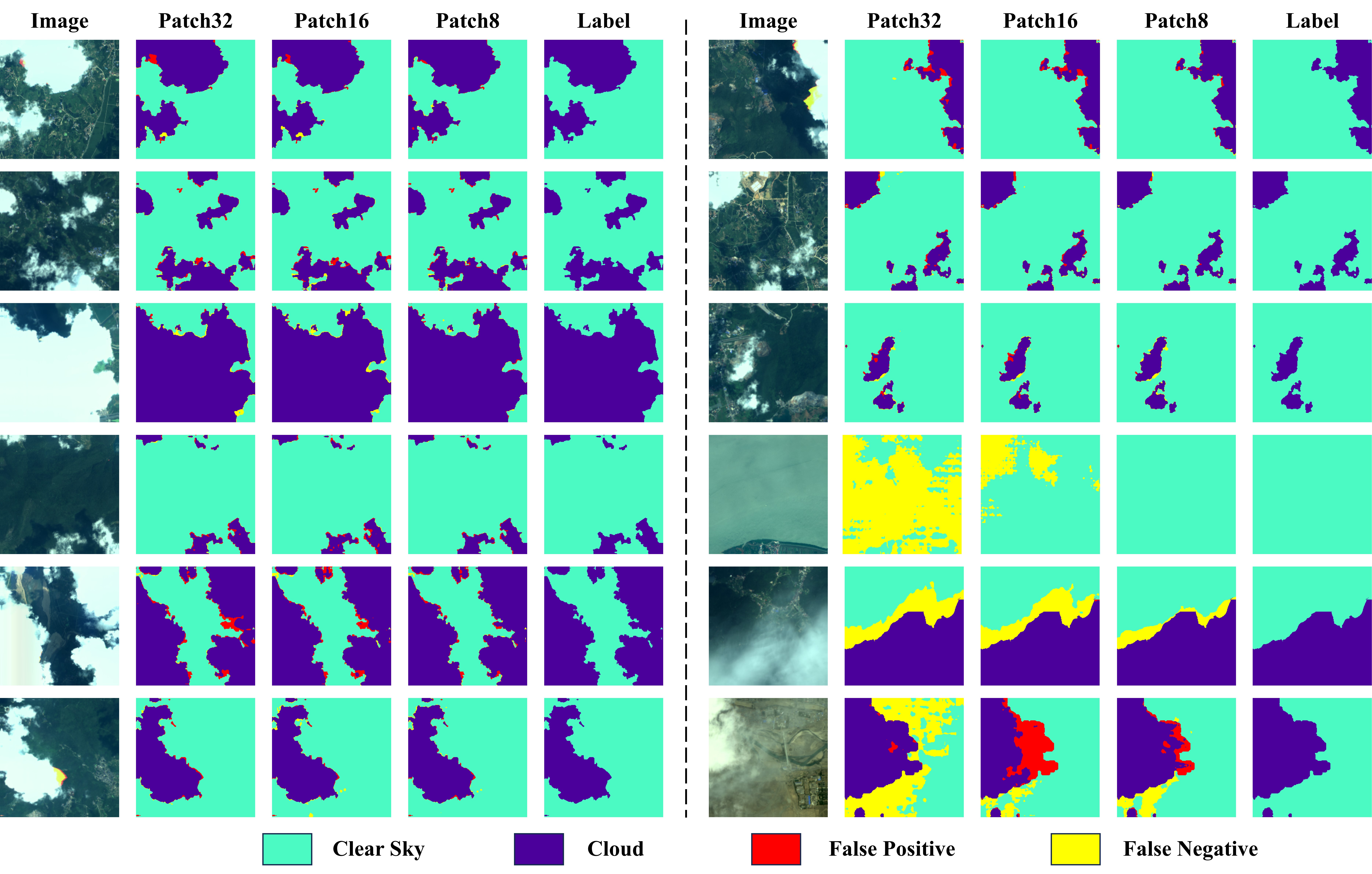} 
	\caption{Comparison of segmentation visualizations from models employing different patch sizes on the GF12MS-WHU-GF1 dataset. As the patch size decreases, the model shows reduced misclassification and omission rates, thereby enhancing its ability to capture fine-grained details in the images.}
	\label{fig:cloud}
\end{figure*}

Due to the large image sizes and limited GPU resources, we set the training batch size to 8, the learning rate to \(1 \times 10^{-4}\), and fine-tuned for 100 epochs. Data augmentation was limited to horizontal and vertical flips applied with a probability of 0.5. Table~\ref{tab:WHU-Building} shows the results of our method compared with several mainstream approaches, including advanced building extraction network frameworks and pre-trained foundational models. The results demonstrate that our approach achieves SOTA performance, with an mIoU of 94.8, which is 1.2 points higher than the second-best method. Moreover, the ablation study presented in Table~\ref{tab:Pixel_level_results} further illustrates the impact of different patch sizes on building extraction accuracy. Figure~\ref{fig:building} presents segmentation visualizations from models employing various patch sizes on the WHU-Building dataset. As the patch size decreases, the model's sensitivity to details—especially building edges—is significantly enhanced, while misclassification and omission rates are reduced.
\begin{table}[!t]
  \centering
  \caption{Fine-Tuning results of different models on the WHU-Building Dataset. Best and second-best results are in red and blue, respectively.}
  \label{tab:WHU-Building}
  \begin{tabular}{lcccc}
    \toprule
    Method      & OA   & Precision & Recall & mIoU \\
    \midrule
    \midrule
    UNet~\cite{ronneberger2015u}        & 97.5 & 94.5      & 94.8   & 89.6 \\
    Deeplabv3~\cite{chen2017rethinking}   & 97.2 & 94.8      & 93.9   & 89.3 \\
    Segformer~\cite{xie2021segformer}   & 97.5 & 94.7      & 94.4   & 89.7 \\
    HRNet~\cite{seong2021semantic}       & 96.8 & 91.7      & 92.9   & 85.6 \\
    BuildFormer~\cite{wang2022building} & 98.0 & 95.2      & 95.1   & 90.7 \\
    RSBuilding~\cite{wang2024rsbuilding}  & 98.3 & 95.9      & 95.8   & 92.2 \\
    DOFA~\cite{xiong2024neural}        & \textbf{\textcolor{blue}{\text{98.7}}} & \textbf{\textcolor{blue}{\text{96.7}}}      & \textbf{\textcolor{blue}{\text{96.6}}}  & \textbf{\textcolor{blue}{\text{93.6}}}\\
    \midrule
    \textbf{FlexiMo (Ours)}  & \textbf{\textcolor{red}{\text{98.9}}}& \textbf{\textcolor{red}{\text{97.2}}}      & \textbf{\textcolor{red}{\text{97.4}}}   & \textbf{\textcolor{red}{\text{94.8}}} \\
    \bottomrule
  \end{tabular}
\end{table}

\subsubsection{Performance on GF12MS-WHU Dataset}
The GF12MS-WHU dataset~\cite{zhu2024transferring} is specifically designed for cloud detection. The data are sourced from Gaofen1-MS images and Gaofen2-MS images. The Gaofen-1 satellite is equipped with two PMS sensors, each containing four MS bands with an 8-meter spatial resolution, along with a panchromatic band at 2-meter resolution. Similarly, the Gaofen-2 satellite features two MS sensors, each with four multispectral bands at a 4-meter spatial resolution, and a panchromatic band at 1-meter resolution. The images in the GF12MS-WHU dataset were collected from various regions in China between June 2014 and December 2020. Each image is sized at $256 \times 256$ pixels, with the dataset comprising 6,434 training images and 4,085 validation images. In this study, we utilize the RGB images from the Gaofen-1 subset for our experiments.
\begin{figure*}[!t]
	\centering
	\includegraphics[width=1.0\linewidth]{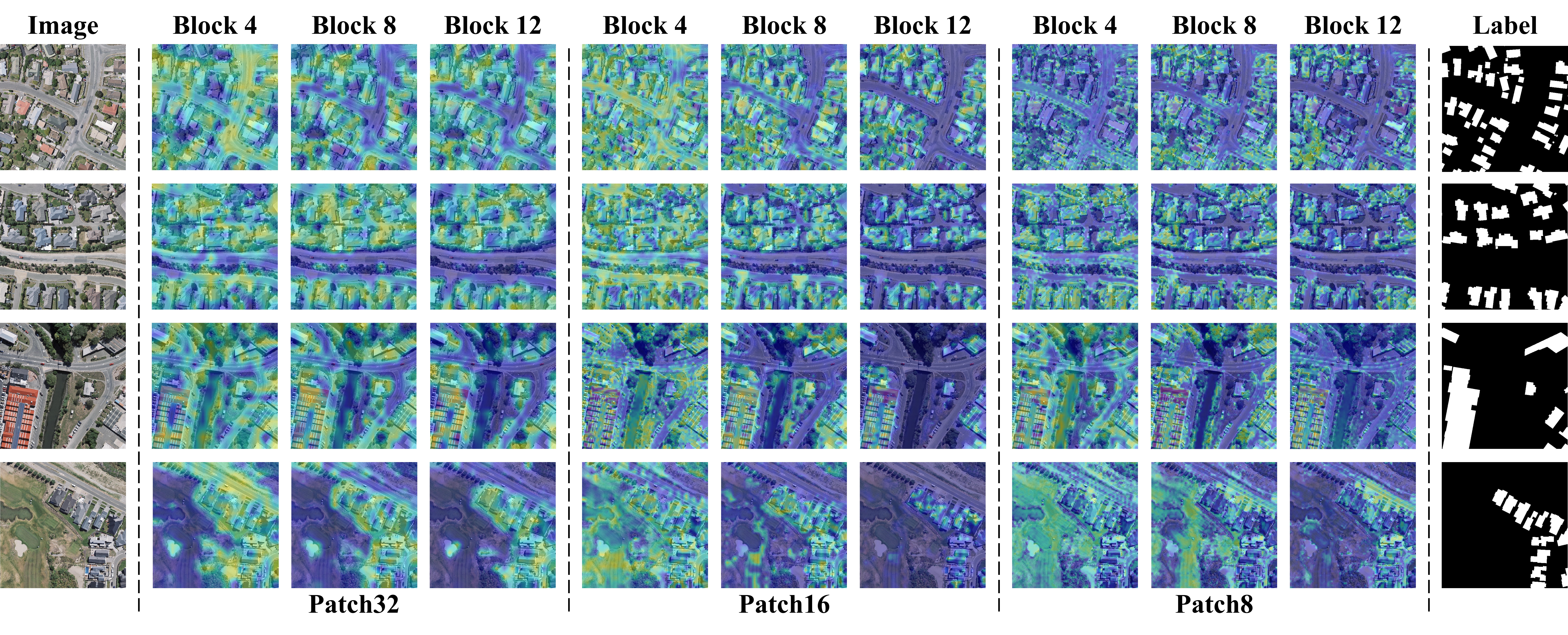} 
	\caption{Visualizations of the model's attention in building extraction tasks. This visualization includes feature maps from various transformer layers in the encoder. Brighter areas indicate higher attention. Moreover, as the patch size decreases, the model's ability to capture fine details and small targets is enhanced.}
	\label{fig:cam_building}
\end{figure*}

Table~\ref{tab:GF12MS} presents the cloud detection results of the proposed method compared to mainstream SOTA approaches. It is evident that our model still achieves the best performance, obtaining an mIoU of 93.8. For fine-tuning, the training configuration is as follows: a batch size of 8, a learning rate of \(4 \times 10^{-5}\), and 100 epochs. The data augmentation strategy remains the same as in previous tasks, employing only horizontal and vertical flips. Compared to DOFA, our model improves the mIoU by 0.6 points, demonstrating the effectiveness of incorporating spatial resolution-aware fine-grained design in downstream fine-tuning of pre-trained foundational models. Figure~\ref{fig:cloud} presents segmentation visualizations from models employing different patch sizes on the GF12MS-WHU-GF1 dataset. As the patch size decreases, the model's misclassification and omission rates are reduced, while its ability to accurately delineate the boundaries of irregular cloud formations is enhanced.

\begin{table}[!t]
  \centering
  \caption{Fine-Tuning results of different models on the GF12MS-WHU-GF1 Dataset. Best and second-best results are in red and blue, respectively.}
  \label{tab:GF12MS}
  \begin{tabular}{lcccc}
    \toprule
    Method         & OA   & Precision & Recall & mIoU \\
    \midrule
    MCDNet~\cite{dong2024mcdnet}         & 97.6 & 91.5      & 92.1   & 85.2 \\
    RSAM-Seg~\cite{zhang2025rsam}       & 98.4 & 94.0      & 94.2   & 89.3 \\
    DBNet~\cite{lu2022dual}          & 98.7 & 96.9      & 93.9   & 91.4 \\
    HRCloudNet~\cite{li2024high}     & 98.8 & 95.0      & 96.3   & 91.9 \\
    SeaMo~\cite{li2024seamomultiseasonalmultimodalremote}          & 98.8 & 96.9      & 94.6   & 92.0 \\
    Cloud-Adapter~\cite{zou2024adapting}  & 98.9 & 95.9      & \textbf{\textcolor{blue}{\text{96.0}}}   & 92.3 \\
    DOFA~\cite{xiong2024neural}           & \textbf{\textcolor{blue}{\text{99.0}}} & \textbf{\textcolor{blue}{\text{96.8}}}      & \textbf{\textcolor{blue}{\text{96.0}}}   & \textbf{\textcolor{blue}{\text{93.2}}} \\
    \midrule
    \textbf{FlexiMo (Ours)}     & \textbf{\textcolor{red}{\text{99.1}}} & \textbf{\textcolor{red}{\text{97.0}}}      & \textbf{\textcolor{red}{\text{96.5}}}   & \textbf{\textcolor{red}{\text{93.8}}} \\
    \bottomrule
  \end{tabular}
\end{table}

\begin{figure}[!t]
	\centering
	\includegraphics[width=0.95\linewidth]{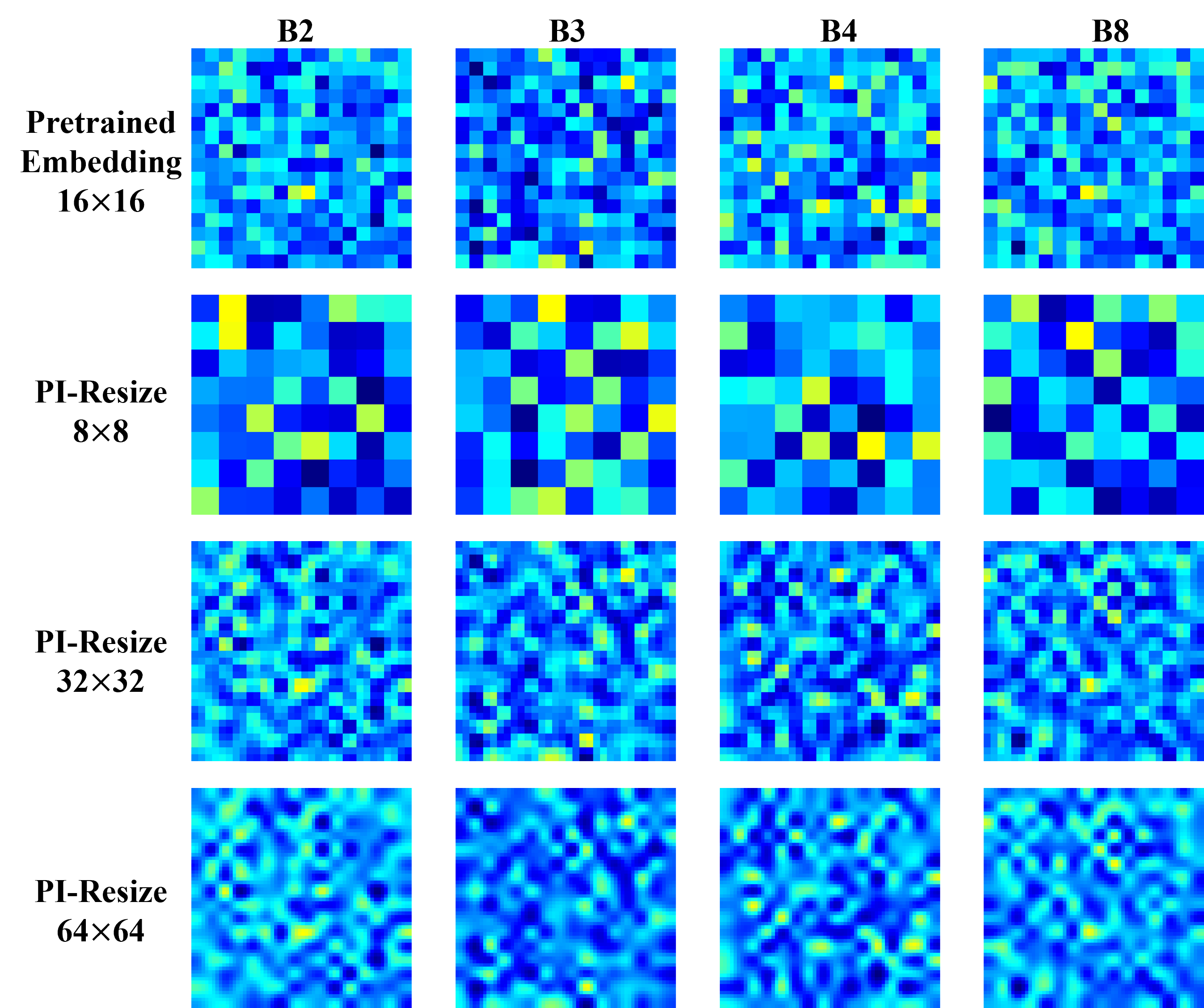} 
	\caption{Visualizations of Spatial Resolution-Aware Module for Sentinel-2 image 10m resolution bands via the parameter-free alignment embedding.}
	\label{fig:conv}
\end{figure}

\subsection{Ablation Studies}
The proposed method in this paper aims to address the challenges posed by multi-size, multi-resolution, multi-channel, and multi-modal remote sensing data within a unified framework. To validate its effectiveness, we conducted ablation studies on various multi-source datasets, adjusting parameters such as image size, patch size, and the number of channels. The results confirm that the proposed method can significantly enhance the performance of pre-trained foundational models on downstream tasks.
\subsubsection{Ablation study on fine-tuning with different image and patch sizes}

\begin{figure*}[!t]
	\centering
	\includegraphics[width=1.0\linewidth]{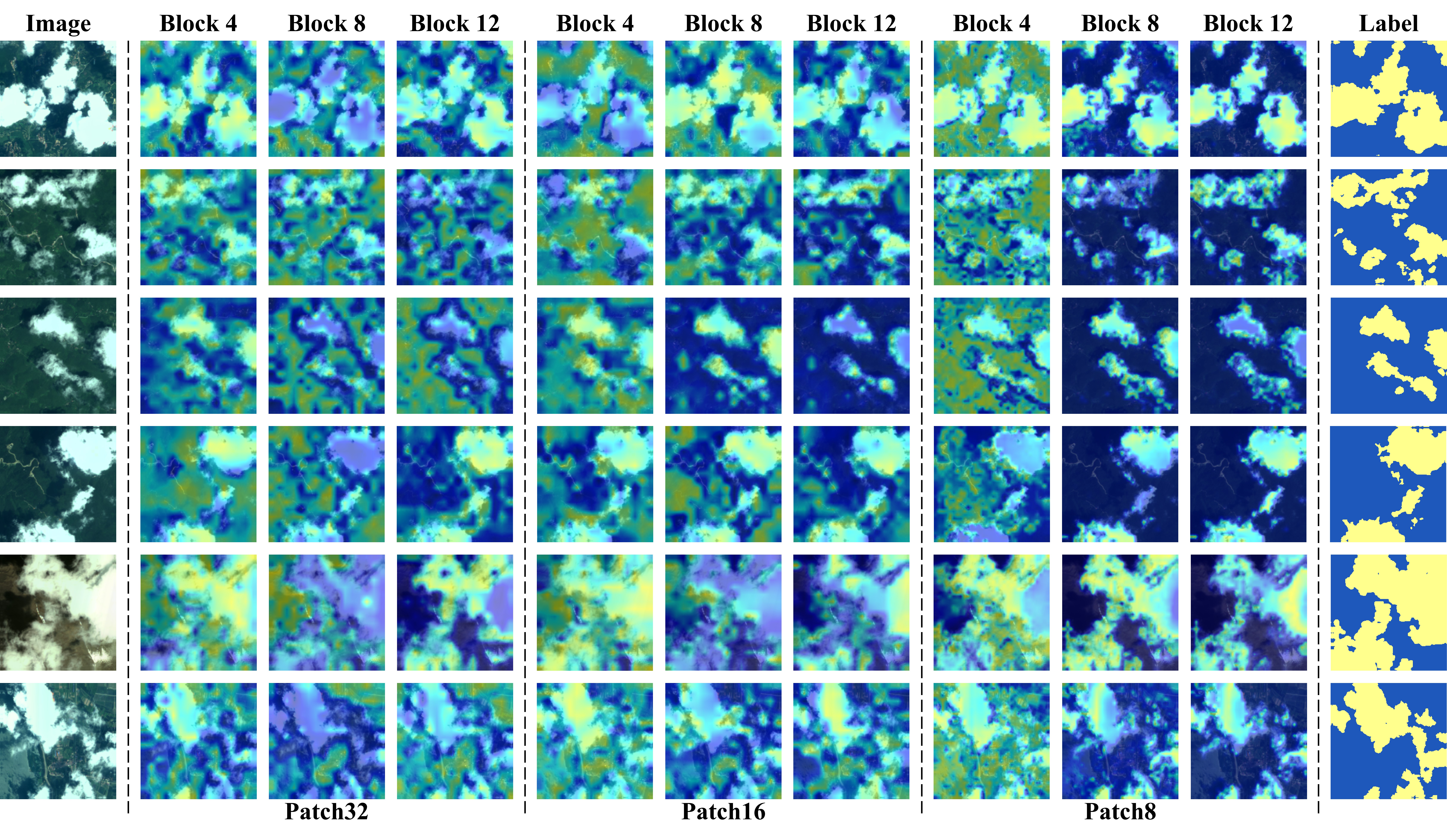} 
	\caption{Visualizations of the model's attention in the cloud detection task. This display includes features from various transformer layers in the encoder. Brighter regions indicate higher levels of model attention.}
	\label{fig:cam_cloud}
\end{figure*}

Table~\ref{tab:tokens_results} presents the performance variations of our proposed method under different image and patch size settings. Specifically, we evaluated five image sizes: \{56, 112, 224, 448, 896\}, and five patch sizes: \{4, 8, 16, 32, 64\}. The results clearly show that, with the incorporation of the spatial awareness module, the fine-tuned pre-trained model maintains robust performance across varying image sizes. 
The experimental results also reveal a trend: for a fixed image size, smaller patch sizes for embedding generally lead to better performance, albeit at the cost of significantly increased token sequence lengths. For instance, when the image size is $896 \times 896$, a patch size of $16 \times 16$ results in a token sequence length of 3,136, whereas a patch size of $64 \times 64$ yields a sequence length of only 196; yet, the accuracy difference between these two settings is a mere 0.07. 
This ablation study indicates that when fine-tuning the pre-trained foundational model (PFM), it is essential to balance model performance and computational cost. In Figure~\ref{fig:conv}, we present visualizations of Sentinel-2 images from different bands (at 10m resolution) after processing through our proposed Spatial Resolution-Aware Module. As the resize dimension decreases, the sensitivity to fine-grained details in the image data increases, which is reflected in the remote sensing imagery as enhanced resolution sensitivity.

\begin{table}[!t]
  \centering
\caption{Ablation Study on the EuroSAT Dataset: Fine-tuning Performance of the Proposed Method with Varying Image and Patch Sizes.}
  \label{tab:tokens_results}
  \begin{tabular}{cccc}
    \toprule
    Image Size      & Patch Size       & Tokens & OA \\
    \midrule
    \midrule
    \multirow{2}{*}{56$\times$56}   & 4$\times$4       & 196    & \textbf{99.43} \\
       & 8$\times$8       & 49     & 99.17 \\
    \midrule
    \multirow{2}{*}{112$\times$112} & 8$\times$8       & 196    & \textbf{99.37} \\
     & 16$\times$16     & 49     & 99.27 \\
    \midrule
    \multirow{2}{*}{224$\times$224} & 8$\times$8       & 784    & \textbf{99.44} \\
     & 16$\times$16     & 196    & 99.39 \\
    \midrule
    \multirow{2}{*}{448$\times$448} & 16$\times$16     & 784    & 99.35 \\
    & 32$\times$32     & 196    & \textbf{99.39} \\
    \midrule
    \multirow{3}{*}{896$\times$896} & 16$\times$16     & 3,136  & \textbf{99.40} \\
     & 32$\times$32     & 784    & 99.38 \\
     & 64$\times$64     & 196    & 99.33 \\
    \bottomrule
  \end{tabular}
\end{table}

\subsubsection{Ablation study on pixel-level fine-tuning tasks with varying patch sizes}
As described in Section~\ref{Performance_on_SegMunich Dataset}, for pixel-level tasks, we do not modify the image size (neither upsampling nor downsampling). Instead, we vary only the patch size over the range of \{4, 8, 16, 32\}. Note that when the patch size is set to $16\times16$, the model is fine-tuned in the same way as the DOFA model, requiring only interpolation of the positional encodings without the need for spatial alignment embedding. Table~\ref{tab:Pixel_level_results} presents the results of our ablation experiments. The findings indicate that, as the granularity of the patch embedding is refined during fine-tuning, the model's performance on dense prediction tasks significantly improves. Moreover, in computationally constrained environments, such as when the patch size is increased to $32\times32$, the performance only exhibits a slight decline. Figure~\ref{fig:cam_building} and Figure~\ref{fig:cam_cloud} present visualizations of the model’s attention across different tasks. These figures display feature maps from various transformer layers in the encoder as well as the saliency maps produced by the segmentation head. The saliency maps were generated using a Class Activation Mapping (CAM) tool~\cite{zhou2016learning}.

\begin{table}[!t]
  \centering
\caption{Ablation Study on the WHU-Building and GF12MS-WHU-GF1 Datasets: Fine-tuning Performance on Pixel-level Tasks with Varying Patch Sizes.}
  \label{tab:Pixel_level_results}
  \begin{tabular}{ccccc}
    \toprule
    Dataset               & Image Size       & Patch Size      & Tokens  & mIoU \\
    \midrule
    \midrule
    \multirow{3}{*}{WHU-Building}        & \multirow{3}{*}{512$\times$512}  & 8$\times$8   & 4,096  & \textbf{94.8} \\
                                          &                                   & 16$\times$16 & 1,024  & 93.6 \\
                                          &                                   & 32$\times$32 & 256    & 93.2 \\
    \midrule
    \multirow{3}{*}{\makecell{GF12MS-WHU-\\GF1}}       & \multirow{3}{*}{256$\times$256}  & 8$\times$8   & 1,024  & \textbf{93.8} \\
                                          &                                   & 16$\times$16 & 256    & 93.2 \\
                                          &                                   & 32$\times$32 & 64     & 93.1 \\
    \midrule
    \multirow{3}{*}{SegMunich} &\multirow{3}{*}{128$\times$128} & 4$\times$4  &1,024& \textbf{52.7} \\
    &          & 8$\times$8     & 256& 50.7 \\ 
                    && 16$\times$16  &64& 47.6 \\
    \bottomrule
  \end{tabular}
\end{table}

\subsubsection{Ablation study on fine-tuning with different channel configurations}
Table~\ref{Channel_Configuration_results} presents the fine-tuning results of our proposed method on different band data within the same sensor type. We conducted ablation experiments on both the EuroSAT MSI and EuroSAT SAR datasets. For the EuroSAT MSI dataset, three sets of channel configurations were selected, each comprising images with different spatial resolutions. For the EuroSAT SAR dataset, two sets of channel configurations were chosen, reflecting different polarization and transmission modes. Although the DOFA model supports arbitrary channel inputs, the experimental results demonstrate that a fixed patch size still constrains the model's ability to extract effective representations. In contrast, our dynamic patch size adjustment improves the performance of the DOFA model across all channel configurations. For example, with 4-band multispectral data at 10\,m resolution, the classification accuracy increased by 1.3\%, while for SAR data with VV and VH channels, the model performance improved from 86.36 to 89.71, corresponding to a 3.35\% boost.
\begin{table}[!t]
  \centering
\caption{Ablation Study on the EuroSAT and EursoSAT-SAR Datasets: Fine-tuning Performance of the Proposed Method with Different Channel Configurations.}
  \label{Channel_Configuration_results}
  \begin{tabular}{cccc}
    \toprule
    Dataset       & Channel                     & Patch Size   & OA \\
    \midrule
    \midrule
    \multirow{6}{*}{EursoSAT} 
                   & \multirow{2}{*}{12 (10m,20m,60m)} & 8$\times$8   & \textbf{99.37} \\
                   &                                  & 16$\times$16 & 99.27 \\
    \cmidrule{2-4}
                   & \multirow{2}{*}{9 (10m,20m)}       & 8$\times$8   & \textbf{99.26} \\
                   &                                  & 16$\times$16 & 99.11 \\
    \cmidrule{2-4}
                   & \multirow{2}{*}{4 (10m)}           & 8$\times$8   & \textbf{99.07} \\
                   &                                  & 16$\times$16 & 98.94 \\
    \midrule
    \multirow{6}{*}{EursoSAT-SAR} 
                   & \multirow{2}{*}{VV,VH}             & 8$\times$8   & \textbf{89.71} \\
                   &                                  & 16$\times$16 & 86.36 \\
    \cmidrule{2-4}
                   & \multirow{2}{*}{VV}                & 8$\times$8   & \textbf{85.18} \\
                   &                                  & 16$\times$16 & 84.35 \\
    \cmidrule{2-4}
                   & \multirow{2}{*}{VH}                & 8$\times$8   & \textbf{86.43} \\
                   &                                  & 16$\times$16 & 85.21 \\
    \bottomrule
  \end{tabular}
\end{table}

\subsubsection{Ablation study on fine-tuning with various resizing strategies}
Since the DOFA model generates patch embeddings through a pre-trained dynamic weight generator module, as described in Section~\ref{overview}, we further process these generated patch embeddings using a Spatial Resolution-Aware Module. To evaluate this processing, we conducted three ablation experiments employing different resizing methods: Linear, MLP, and PI-Resize. The \emph{Linear} method simply applies linear interpolation to the generated convolutional kernels. The \emph{MLP} method inserts a fully connected layer to transform the weights to the required shape after the DOFA-generated weights but before reshaping them into convolutional kernels. The third method, \emph{PI-Resize}, is the approach we propose. Table~\ref{tab:Albation_resizing_methods} presents the results of these three resizing methods on the EuroSAT MSI dataset. The results indicate that only PI-Resize can maintain or even slightly improve model performance when the patch size is increased, whereas the other two methods lead to a decline in performance.
\begin{table}[!t]
\centering
\caption{Ablation Study on the EuroSAT Dataset: Fine-tuning Performance of the Proposed Method Using Different Resizing Strategies.}
\label{tab:Albation_resizing_methods}
\begin{tabular}{ccccc}
\toprule
Image Size & Patch Size & Method     &OA  \\
\midrule
\midrule
448x448    & 16x16      & Fixed      & 99.35                         \\
\midrule
448x448    & 32x32      & Linear     & 99.12                      \\
448x448    & 32x32      & MLP        & 98.76                      \\
448x448    & 32x32      & PI-Resize  & \textbf{99.39}                    \\
\bottomrule
\end{tabular}
\end{table}

\section{Conclusion}
In this study, we introduced FlexiMo, a flexible remote sensing foundation model designed to address the challenges posed by the heterogeneous nature of remote sensing imagery. FlexiMo enhances spatial adaptability by dynamically adjusting patch embeddings to accommodate varying spatial resolutions and patch sizes. At its core, a novel spatial resolution-aware module employs a resolution alignment strategy to recalibrate patch embeddings, preserving critical token characteristics and ensuring multi-scale feature fidelity—all without modifying the underlying network architecture. This design enables seamless processing of diverse remote sensing data while maintaining the integrity of pre-trained models.

To assess FlexiMo's effectiveness, we conducted extensive experiments across multiple multimodal, multi-resolution, and multi-scale datasets. Our results consistently demonstrate that FlexiMo outperforms existing methods across all benchmarks. Additionally, ablation studies confirm its robustness and flexibility under varying image sizes, patch sizes, and channel configurations. FlexiMo represents a significant advancement in remote sensing image analysis. By effectively addressing the challenges of spatial adaptation, it paves the way for more generalizable and efficient foundation models capable of adapting to diverse spatial conditions.

\bibliographystyle{ieeetr}
\bibliography{my_ref}

\begin{thebibliography}{10}

\bibitem{zhao2024artificial}
T.~Zhao, S.~Wang, C.~Ouyang, M.~Chen, C.~Liu, J.~Zhang, L.~Yu, F.~Wang, Y.~Xie, J.~Li, {\em et~al.}, ``Artificial intelligence for geoscience: Progress, challenges and perspectives,'' {\em The Innovation}, 2024.

\bibitem{Li2024Interpretable}
C.~Li, D.~Hong, B.~Zhang, T.~Liao, N.~Yokoya, P.~Ghamisi, M.~Chen, L.~Wang, J.~A. Benediktsson, and J.~Chanussot, ``Interpretable foundation models as decryptors peering into the earth system,'' {\em The Innovation}, vol.~5, no.~5, p.~100682, 2024.

\bibitem{cong2022satmae}
Y.~Cong, S.~Khanna, C.~Meng, P.~Liu, E.~Rozi, Y.~He, M.~Burke, D.~Lobell, and S.~Ermon, ``Satmae: Pre-training transformers for temporal and multi-spectral satellite imagery,'' {\em Advances in Neural Information Processing Systems}, vol.~35, pp.~197--211, 2022.

\bibitem{li2024s2mae}
X.~Li, D.~Hong, and J.~Chanussot, ``S2mae: A spatial-spectral pretraining foundation model for spectral remote sensing data,'' in {\em Proceedings of the IEEE/CVF Conference on Computer Vision and Pattern Recognition}, pp.~24088--24097, 2024.

\bibitem{hong2024multimodal}
D.~Hong, C.~Li, B.~Zhang, N.~Yokoya, J.~A. Benediktsson, and J.~Chanussot, ``Multimodal artificial intelligence foundation models: Unleashing the power of remote sensing big data in earth observation,'' {\em The Innovation Geoscience}, vol.~2, no.~1, p.~100055, 2024.

\bibitem{li2024seamomultiseasonalmultimodalremote}
X.~Li, D.~Hong, C.~Li, and J.~Chanussot, ``Seamo: A multi-seasonal and multimodal remote sensing foundation model,'' 2024.

\bibitem{hong2024spectralgpt}
D.~Hong, B.~Zhang, X.~Li, Y.~Li, C.~Li, J.~Yao, N.~Yokoya, H.~Li, P.~Ghamisi, X.~Jia, {\em et~al.}, ``Spectralgpt: Spectral remote sensing foundation model,'' {\em IEEE Transactions on Pattern Analysis and Machine Intelligence}, 2024.

\bibitem{10504785}
F.~Liu, D.~Chen, Z.~Guan, X.~Zhou, J.~Zhu, Q.~Ye, L.~Fu, and J.~Zhou, ``Remoteclip: A vision language foundation model for remote sensing,'' {\em IEEE Transactions on Geoscience and Remote Sensing}, vol.~62, pp.~1--16, 2024.

\bibitem{kuckreja2023geochat}
K.~Kuckreja, M.~S. Danish, M.~Naseer, A.~Das, S.~Khan, and F.~S. Khan, ``Geochat: Grounded large vision-language model for remote sensing,'' {\em The IEEE/CVF Conference on Computer Vision and Pattern Recognition}, 2024.

\bibitem{astruc2024anysat}
G.~Astruc, N.~Gonthier, C.~Mallet, and L.~Landrieu, ``Anysat: An earth observation model for any resolutions, scales, and modalities,'' {\em arXiv preprint arXiv:2412.14123}, 2024.

\bibitem{reed2023scale}
C.~J. Reed, R.~Gupta, S.~Li, S.~Brockman, C.~Funk, B.~Clipp, K.~Keutzer, S.~Candido, M.~Uyttendaele, and T.~Darrell, ``Scale-mae: A scale-aware masked autoencoder for multiscale geospatial representation learning,'' in {\em Proceedings of the IEEE/CVF International Conference on Computer Vision}, pp.~4088--4099, 2023.

\bibitem{prexl2024senpa}
J.~Prexl and M.~Schmitt, ``Senpa-mae: Sensor parameter aware masked autoencoder for multi-satellite self-supervised pretraining,'' {\em arXiv preprint arXiv:2408.11000}, 2024.

\bibitem{xiong2024neural}
Z.~Xiong, Y.~Wang, F.~Zhang, A.~J. Stewart, J.~Hanna, D.~Borth, I.~Papoutsis, B.~L. Saux, G.~Camps-Valls, and X.~X. Zhu, ``Neural plasticity-inspired multimodal foundation model for earth observation,'' {\em arXiv preprint arXiv:2403.15356}, 2024.

\bibitem{dosovitskiy2021an}
A.~Dosovitskiy, L.~Beyer, A.~Kolesnikov, D.~Weissenborn, X.~Zhai, T.~Unterthiner, M.~Dehghani, M.~Minderer, G.~Heigold, S.~Gelly, J.~Uszkoreit, and N.~Houlsby, ``An image is worth 16x16 words: Transformers for image recognition at scale,'' in {\em International Conference on Learning Representations}, 2021.

\bibitem{wang2025scaling}
F.~Wang, Y.~Yu, G.~Wei, W.~Shao, Y.~Zhou, A.~Yuille, and C.~Xie, ``Scaling laws in patchification: An image is worth 50,176 tokens and more,'' {\em arXiv preprint arXiv:2502.03738}, 2025.

\bibitem{liu2021swin}
Z.~Liu, Y.~Lin, Y.~Cao, H.~Hu, Y.~Wei, Z.~Zhang, S.~Lin, and B.~Guo, ``Swin transformer: Hierarchical vision transformer using shifted windows,'' in {\em Proceedings of the IEEE/CVF international conference on computer vision}, pp.~10012--10022, 2021.

\bibitem{katharopoulos2020transformers}
A.~Katharopoulos, A.~Vyas, N.~Pappas, and F.~Fleuret, ``Transformers are rnns: Fast autoregressive transformers with linear attention,'' in {\em International conference on machine learning}, pp.~5156--5165, PMLR, 2020.

\bibitem{liu2023efficientvit}
X.~Liu, H.~Peng, N.~Zheng, Y.~Yang, H.~Hu, and Y.~Yuan, ``Efficientvit: Memory efficient vision transformer with cascaded group attention,'' in {\em Proceedings of the IEEE/CVF conference on computer vision and pattern recognition}, pp.~14420--14430, 2023.

\bibitem{beyer2023flexivit}
L.~Beyer, P.~Izmailov, A.~Kolesnikov, M.~Caron, S.~Kornblith, X.~Zhai, M.~Minderer, M.~Tschannen, I.~Alabdulmohsin, and F.~Pavetic, ``Flexivit: One model for all patch sizes,'' in {\em Proceedings of the IEEE/CVF Conference on Computer Vision and Pattern Recognition}, pp.~14496--14506, 2023.

\bibitem{fan2024vitar}
Q.~Fan, Q.~You, X.~Han, Y.~Liu, Y.~Tao, H.~Huang, R.~He, and H.~Yang, ``Vitar: Vision transformer with any resolution,'' {\em arXiv preprint arXiv:2403.18361}, 2024.

\bibitem{liumspe}
W.~Liu, F.~Zhu, S.~Ma, and C.-L. Liu, ``Mspe: Multi-scale patch embedding prompts vision transformers to any resolution,'' in {\em The Thirty-eighth Annual Conference on Neural Information Processing Systems}, 2024.

\bibitem{irvin2023usat}
J.~Irvin, L.~Tao, J.~Zhou, Y.~Ma, L.~Nashold, B.~Liu, and A.~Y. Ng, ``Usat: A unified self-supervised encoder for multi-sensor satellite imagery,'' {\em arXiv preprint arXiv:2312.02199}, 2023.

\bibitem{he2022masked}
K.~He, X.~Chen, S.~Xie, Y.~Li, P.~Doll{\'a}r, and R.~Girshick, ``Masked autoencoders are scalable vision learners,'' in {\em Proceedings of the IEEE/CVF conference on computer vision and pattern recognition}, pp.~16000--16009, 2022.

\bibitem{fuller2023croma}
A.~Fuller, K.~Millard, and J.~R. Green, ``Croma: Remote sensing representations with contrastive radar-optical masked autoencoders,'' in {\em Thirty-seventh Conference on Neural Information Processing Systems}, 2023.

\bibitem{ayush2021geography}
K.~Ayush, B.~Uzkent, C.~Meng, K.~Tanmay, M.~Burke, D.~Lobell, and S.~Ermon, ``Geography-aware self-supervised learning,'' in {\em Proceedings of the IEEE/CVF International Conference on Computer Vision}, pp.~10181--10190, 2021.

\bibitem{10655854}
X.~Guo, J.~Lao, B.~Dang, Y.~Zhang, L.~Yu, L.~Ru, L.~Zhong, Z.~Huang, K.~Wu, D.~Hu, H.~He, J.~Wang, J.~Chen, M.~Yang, Y.~Zhang, and Y.~Li, ``Skysense: A multi-modal remote sensing foundation model towards universal interpretation for earth observation imagery,'' in {\em 2024 IEEE/CVF Conference on Computer Vision and Pattern Recognition (CVPR)}, pp.~27662--27673, 2024.

\bibitem{zhang2024uv}
X.~Zhang, Y.~Liu, Y.~Lin, Q.~Liao, and Y.~Li, ``Uv-sam: Adapting segment anything model for urban village identification,'' in {\em Proceedings of the AAAI Conference on Artificial Intelligence}, vol.~38, pp.~22520--22528, 2024.

\bibitem{li2025urbansam}
C.~Li, D.~Hong, B.~Zhang, Y.~Li, G.~Camps-Valls, X.~X. Zhu, and J.~Chanussot, ``Urbansam: Learning invariance-inspired adapters for segment anything models in urban construction,'' {\em arXiv preprint arXiv:2502.15199}, 2025.

\bibitem{10663449}
D.~Tang, X.~Cao, X.~Hou, Z.~Jiang, J.~Liu, and D.~Meng, ``Crs-diff: Controllable remote sensing image generation with diffusion model,'' {\em IEEE Transactions on Geoscience and Remote Sensing}, vol.~62, pp.~1--14, 2024.

\bibitem{moore1920reciprocal}
E.~H. Moore, ``On the reciprocal of the general algebraic matrix,'' {\em Bulletin of the american mathematical society}, vol.~26, pp.~294--295, 1920.

\bibitem{helber2019eurosat}
P.~Helber, B.~Bischke, A.~Dengel, and D.~Borth, ``Eurosat: A novel dataset and deep learning benchmark for land use and land cover classification,'' {\em IEEE Journal of Selected Topics in Applied Earth Observations and Remote Sensing}, vol.~12, no.~7, pp.~2217--2226, 2019.

\bibitem{wang2023feature}
Y.~Wang, H.~H. Hern{\'a}ndez, C.~M. Albrecht, and X.~X. Zhu, ``Feature guided masked autoencoder for self-supervised learning in remote sensing,'' {\em arXiv preprint arXiv:2310.18653}, 2023.

\bibitem{ji2018fully}
S.~Ji, S.~Wei, and M.~Lu, ``Fully convolutional networks for multisource building extraction from an open aerial and satellite imagery data set,'' {\em IEEE Transactions on geoscience and remote sensing}, vol.~57, no.~1, pp.~574--586, 2018.

\bibitem{zhu2024transferring}
S.~Zhu, Z.~Li, and H.~Shen, ``Transferring deep models for cloud detection in multisensor images via weakly supervised learning,'' {\em IEEE Transactions on Geoscience and Remote Sensing}, vol.~62, pp.~1--18, 2024.

\bibitem{xiao2018unified}
T.~Xiao, Y.~Liu, B.~Zhou, Y.~Jiang, and J.~Sun, ``Unified perceptual parsing for scene understanding,'' in {\em Proceedings of the European conference on computer vision (ECCV)}, pp.~418--434, 2018.

\bibitem{manas2021seasonal}
O.~Manas, A.~Lacoste, X.~Gir{\'o}-i Nieto, D.~Vazquez, and P.~Rodriguez, ``Seasonal contrast: Unsupervised pre-training from uncurated remote sensing data,'' in {\em Proceedings of the IEEE/CVF International Conference on Computer Vision}, pp.~9414--9423, 2021.

\bibitem{wang2022self}
Y.~Wang, C.~M. Albrecht, and X.~X. Zhu, ``Self-supervised vision transformers for joint sar-optical representation learning,'' in {\em IGARSS 2022-2022 IEEE International Geoscience and Remote Sensing Symposium}, pp.~139--142, IEEE, 2022.

\bibitem{ronneberger2015u}
O.~Ronneberger, P.~Fischer, and T.~Brox, ``U-net: Convolutional networks for biomedical image segmentation,'' in {\em Medical image computing and computer-assisted intervention--MICCAI 2015: 18th international conference, Munich, Germany, October 5-9, 2015, proceedings, part III 18}, pp.~234--241, Springer, 2015.

\bibitem{chen2017rethinking}
L.-C. Chen, G.~Papandreou, F.~Schroff, and H.~Adam, ``Rethinking atrous convolution for semantic image segmentation,'' {\em arXiv preprint arXiv:1706.05587}, 2017.

\bibitem{xie2021segformer}
E.~Xie, W.~Wang, Z.~Yu, A.~Anandkumar, J.~M. Alvarez, and P.~Luo, ``Segformer: Simple and efficient design for semantic segmentation with transformers,'' {\em Advances in neural information processing systems}, vol.~34, pp.~12077--12090, 2021.

\bibitem{seong2021semantic}
S.~Seong and J.~Choi, ``Semantic segmentation of urban buildings using a high-resolution network (hrnet) with channel and spatial attention gates,'' {\em Remote Sensing}, vol.~13, no.~16, p.~3087, 2021.

\bibitem{wang2022building}
L.~Wang, S.~Fang, X.~Meng, and R.~Li, ``Building extraction with vision transformer,'' {\em IEEE Transactions on Geoscience and Remote Sensing}, vol.~60, pp.~1--11, 2022.

\bibitem{wang2024rsbuilding}
M.~Wang, L.~Su, C.~Yan, S.~Xu, P.~Yuan, X.~Jiang, and B.~Zhang, ``Rsbuilding: towards general remote sensing image building extraction and change detection with foundation model,'' {\em IEEE Transactions on Geoscience and Remote Sensing}, 2024.

\bibitem{dong2024mcdnet}
J.~Dong, Y.~Wang, Y.~Yang, M.~Yang, and J.~Chen, ``Mcdnet: Multilevel cloud detection network for remote sensing images based on dual-perspective change-guided and multi-scale feature fusion,'' {\em International Journal of Applied Earth Observation and Geoinformation}, vol.~129, p.~103820, 2024.

\bibitem{zhang2025rsam}
J.~Zhang, Y.~Li, X.~Yang, R.~Jiang, and L.~Zhang, ``Rsam-seg: A sam-based model with prior knowledge integration for remote sensing image semantic segmentation,'' {\em Remote Sensing}, vol.~17, no.~4, p.~590, 2025.

\bibitem{lu2022dual}
C.~Lu, M.~Xia, M.~Qian, and B.~Chen, ``Dual-branch network for cloud and cloud shadow segmentation,'' {\em IEEE Transactions on Geoscience and Remote Sensing}, vol.~60, pp.~1--12, 2022.

\bibitem{li2024high}
J.~Li, T.~Xue, J.~Zhao, J.~Ge, Y.~Min, W.~Su, and K.~Zhan, ``High-resolution cloud detection network,'' {\em Journal of Electronic Imaging}, vol.~33, no.~4, pp.~043027--043027, 2024.

\bibitem{zou2024adapting}
X.~Zou, S.~Zhang, K.~Li, S.~Wang, J.~Xing, L.~Jin, C.~Lang, and P.~Tao, ``Adapting vision foundation models for robust cloud segmentation in remote sensing images,'' {\em arXiv preprint arXiv:2411.13127}, 2024.

\bibitem{zhou2016learning}
B.~Zhou, A.~Khosla, A.~Lapedriza, A.~Oliva, and A.~Torralba, ``Learning deep features for discriminative localization,'' in {\em Proceedings of the IEEE conference on computer vision and pattern recognition}, pp.~2921--2929, 2016.

\end{thebibliography}

\end{document}